\documentclass{article}

\usepackage{iclr2026_conference,times}
\usepackage[T1]{fontenc}
\usepackage[utf8]{inputenc}
\usepackage{wrapfig}
\usepackage{amsmath,amssymb}
\usepackage{graphicx}
\usepackage{booktabs}
\usepackage{multirow}
\usepackage{tabularx}
\usepackage{placeins}
\usepackage{needspace}
\usepackage{float}
\usepackage{capt-of}
\usepackage{microtype}
\usepackage{fancyhdr}
\usepackage{xcolor}
\usepackage[skins,breakable]{tcolorbox}
\usepackage{caption}
\usepackage{hyperref}
\usepackage{url}
\usepackage{fontawesome}
\input{appendix/grading_prompt_style}
\definecolor{templateblue}{RGB}{40,78,116}
\definecolor{mintblue}{RGB}{224,240,245}
\definecolor{mintframe}{RGB}{161,185,199}
\definecolor{minttitle}{RGB}{56,100,125}
\newtcolorbox{casebox}[1]{
    enhanced,
    breakable,
    colback=mintblue!40!white,
    colframe=mintframe,
    colbacktitle=minttitle!15!white,
    coltitle=templateblue,
    title=\textbf{#1},
    fonttitle=\bfseries,
    sharp corners=south,
    borderline={0.8pt}{0pt}{minttitle},
    boxrule=0.8pt,
    arc=6pt,
    left=6pt, right=6pt, top=6pt, bottom=6pt,
    before upper={\setlength{\parskip}{1pt}},
    before skip=6pt, after skip=6pt,
    drop shadow=black!15
}
\hypersetup{
    colorlinks=true,
    linkcolor=templateblue,
    citecolor=templateblue,
    urlcolor=templateblue,
    pdftitle={From Research Goals to Scientific Closure: Evaluating AI Agents Through Evidence-Supported Physics Research},
    pdfauthor={Anonymous Authors}
}
\newcommand{\xmark}{\ensuremath{\times}}

\graphicspath{{figures/}}


\title{
\vspace{-6mm}
From Research Goals to Scientific Closure:
Evaluating \\ Agents Through 
Evidence-Supported Physics Research
}

\author{%
\begin{minipage}{0.96\textwidth}
\vspace*{0.2em}
\centering
\normalfont
{\small\textbf{Yijia Mao}\textsuperscript{1,2*} \enspace
\textbf{Song Dai}\textsuperscript{1,3*}\enspace 
\textbf{Tao Chen}\textsuperscript{2}\enspace 
\textbf{Xiangyu Lu}\textsuperscript{2}\enspace
\textbf{Linqian Zeng}\textsuperscript{2}\enspace 
\textbf{Chenxi Hu}\textsuperscript{2}\enspace \\
\textbf{Sitong Zhou}\textsuperscript{2}\enspace
\textbf{Dongyang He}\textsuperscript{2}\enspace
\textbf{Xiaodan Mao}\textsuperscript{2}\enspace
\textbf{Linxuan Zhang}\textsuperscript{2}\enspace        
\textbf{Tong Zhu}\textsuperscript{1}\enspace 
\\
\textbf{Yafu Li}\textsuperscript{1}\enspace
\textbf{Fangchen Yu}\textsuperscript{1$\ddagger$}\enspace  
\textbf{Peng Ye}\textsuperscript{1$\ddagger$}\enspace
\textbf{Feng He}\textsuperscript{2}\enspace
\textbf{Yu Cheng}\textsuperscript{4$\ddagger$}\enspace
\textbf{Yun Luo}\textsuperscript{1\textdagger,$\ddagger$}\enspace
}\\[0.7em]
{\footnotesize
\textsuperscript{1}\,Shanghai AI Laboratory \quad
\textsuperscript{2}\,Shanghai Jiao Tong University \\[0.3em]
\textsuperscript{3}\,The HongKong University of Science and Technology(Guangzhou) \\[0.3em]
\textsuperscript{4} Nanyang Technological University
\\[0.3em]
\textsuperscript{*}Equal Contribution,$\quad$\textsuperscript{\textdagger}Project Lead,$\quad$
\textsuperscript{$\ddagger$}Corresponding Author \\
}
\end{minipage}%
}

\iclrfinalcopy

\begin{document}

\maketitle
\begingroup
\renewcommand{\thefootnote}{}
\footnotetext{Github Repository: \url{https://github.com/uncertainty2019/PhysClosureBench}}
\endgroup
\pagestyle{fancy}
\fancyhf{}
\fancyhead[L]{%
  \raisebox{0pt}[0pt][0pt]{%
    \footnotesize
    From Research Goals to Scientific Closure:
    Evaluating Agents Through Evidence-Supported Physics Research
  }%
}
\renewcommand{\headrulewidth}{0.4pt}


\vspace{-6mm}
\begin{abstract}
\vspace{-3mm}
Can AI agents organize scientific investigations that lead to well-supported conclusions?
To evaluate this capability, we introduce \textbf{\textit{PhysClosureBench}}, a benchmark of 50 expert-verified, goal-specified research tasks spanning strong-field, attosecond, and neighboring physics areas.
Each task specifies the research objective and essential scientific constraints, while leaving the agent to organize the investigation and determine how to generate and validate the evidence needed to answer the question.
A 20-criterion Atomic Score measures local scientific accomplishment, while Research Closure (RC0--RC4) assesses how fully the investigation resolves the specified research question based on whether its indispensable evidence requirements and applicable scope conditions are satisfied.
We evaluate {seven} frontier LLMs, where
Claude-Opus-5 achieves the highest mean Atomic Score (6.29/10) but reaches RC4 in only 32.7\% of attempts.
Across the benchmark, 36.2\% of attempts scoring at least 8/10 still fall short of RC4, while all seven models reach RC4 on substantially fewer tasks in all three runs than in at least one.
Case-level analysis further illustrates how misinterpretation of generated results and unresolved physical applicability limits can lead to incomplete closure.
Together, these results reveal a persistent gap between AI agents' local scientific accomplishments and their ability to achieve complete research closure.

\end{abstract}

\vspace{-3mm}

\providecommand{\NTasks}{50}
\providecommand{\NModels}{7}
\providecommand{\NRunsPerCase}{3}
\providecommand{\NRegisteredEvaluations}{1050}
\providecommand{\NRubricAtoms}{20}
\providecommand{\NPhysicsDomains}{5}
\providecommand{\MeanScoreGPT}{5.83}
\providecommand{\MeanScoreKimi}{5.17}
\providecommand{\MeanScoreDeepSeek}{4.50}
\providecommand{\MeanScoreQwen}{3.46}
\providecommand{\MeanScoreGLM}{3.28}
\providecommand{\MeanScoreGemini}{2.85}
\providecommand{\HighScoreRCFourPercent}{63.8}
\providecommand{\HighScoreNonRCFourPercent}{36.2}
\providecommand{\RCFourProfileF}{99.5}
\providecommand{\RCFourProfileO}{97.1}
\providecommand{\RCFourProfileI}{97.5}
\providecommand{\RCFourProfileM}{96.8}
\providecommand{\NonRCFourProfileF}{66.0}
\providecommand{\NonRCFourProfileO}{31.1}
\providecommand{\NonRCFourProfileI}{23.8}
\providecommand{\NonRCFourProfileM}{24.0}
\providecommand{\AtLeastOneRCFourGPT}{40.0}
\providecommand{\AllRCFourGPT}{10.0}
\providecommand{\AtLeastOneRCFourKimi}{40.0}
\providecommand{\AllRCFourKimi}{4.0}
\providecommand{\AtLeastOneRCFourDeepSeek}{24.0}
\providecommand{\AllRCFourDeepSeek}{0.0}
\providecommand{\AtLeastOneRCFourQwen}{14.0}
\providecommand{\AllRCFourQwen}{2.0}
\providecommand{\AtLeastOneRCFourGLM}{12.0}
\providecommand{\AllRCFourGLM}{0.0}
\providecommand{\AtLeastOneRCFourGemini}{2.0}
\providecommand{\AllRCFourGemini}{0.0}
\providecommand{\MeanScoreOpus}{6.29}
\providecommand{\AtLeastOneRCFourOpus}{50.0}
\providecommand{\AllRCFourOpus}{18.0}
\providecommand{\NFinalAttempts}{1050}
\providecommand{\NAttemptsPerModel}{150}
\providecommand{\NRCFourAttempts}{148}
\providecommand{\NNonRCFourAttempts}{902}
\providecommand{\VScopeSatisfied}{852}
\providecommand{\VScopeApplicable}{981}
\providecommand{\VScopePercent}{86.9}
\providecommand{\VRequiredValidationSatisfied}{93}
\providecommand{\VRequiredValidationApplicable}{156}
\providecommand{\VRequiredValidationPercent}{59.6}
\providecommand{\VRiskLinkedValidationSatisfied}{441}
\providecommand{\VRiskLinkedValidationApplicable}{922}
\providecommand{\VRiskLinkedValidationPercent}{47.8}
\providecommand{\RCFourPercentOpus}{32.7}
\providecommand{\MedianScoreOpus}{6.75}
\providecommand{\MeanScoreOverall}{4.48}
\providecommand{\OverallRCFourPercent}{14.1}
\providecommand{\VAllSatisfied}{1386}
\providecommand{\VAllApplicable}{2059}
\providecommand{\NScoreAtLeastFour}{525}
\providecommand{\NRCFourScoreAtLeastFour}{148}
\providecommand{\NScoreAtLeastSix}{374}
\providecommand{\NRCFourScoreAtLeastSix}{148}
\providecommand{\NScoreAtLeastEight}{229}
\providecommand{\NRCFourScoreAtLeastEight}{146}

\begin{figure}[!htbp]
    \centering
    \includegraphics[width=0.95\linewidth]{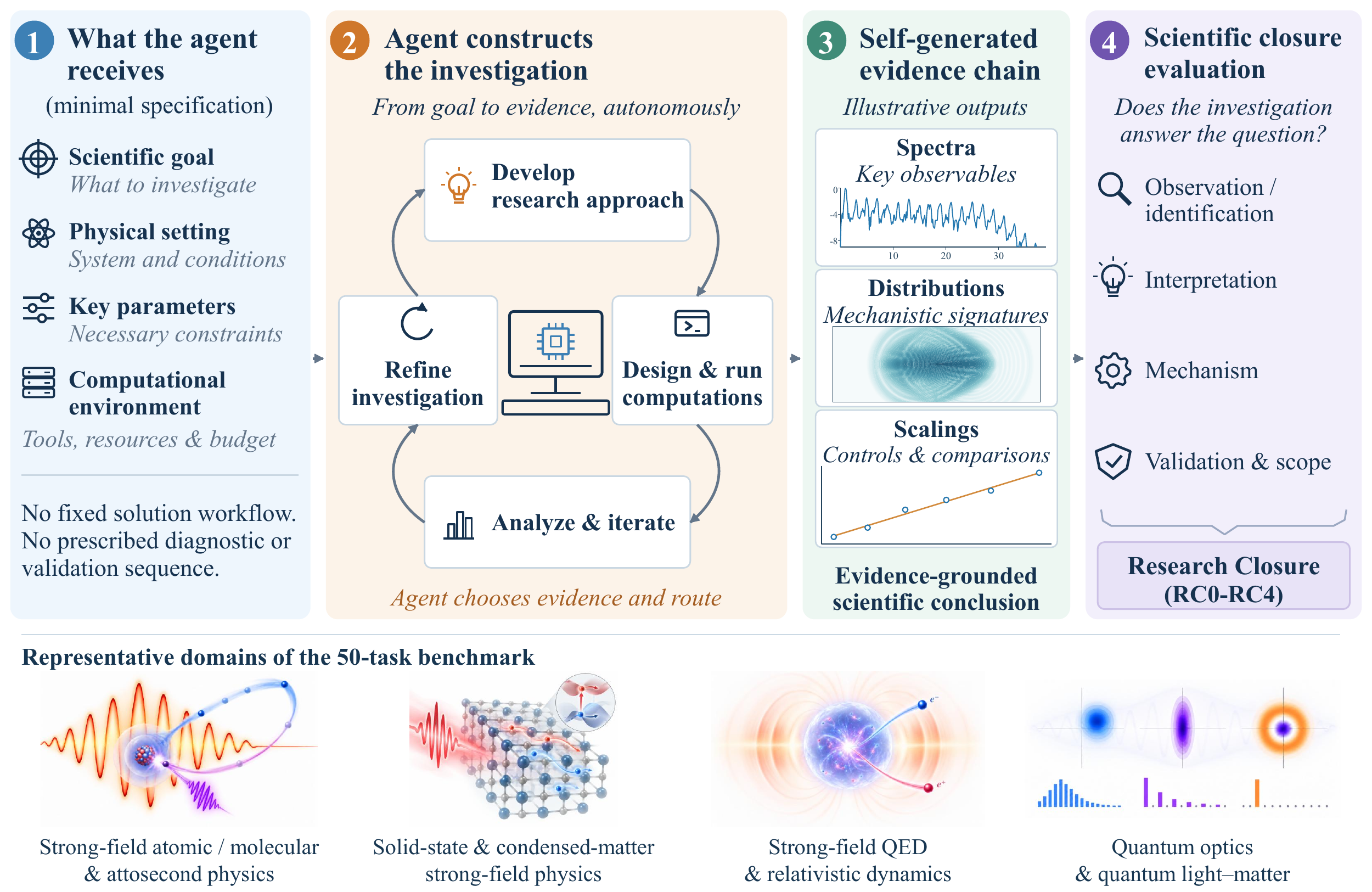}
    \caption{
    \textbf{From a scientific goal to scientific closure.}
    In PhysClosureBench, agents are required to achieve the provided scientific objective within the specified task setting and a resource-bounded computational environment. 
    }
    \label{fig:benchmark_overview}
\end{figure}

\section{Introduction}
\label{sec:introduction}

\begin{quote}
\small\itshape
``Science is built up of facts, as a house is built of stones; but an
accumulation of facts is no more a science than a heap of stones is a house.''
\par\hfill\upshape--- Henri Poincaré.
\end{quote}

As language-model agents become more capable of reasoning, writing code, and processing long contexts, they are increasingly applied to scientific research, assisting with literature synthesis, scientific programming, and the reproduction of prior work \citep{lu2024aiscientist,starace2025paperbench,wijk2025rebench,zheng2025autonomy,zhang2024scientificllms}. While these capabilities make them promising scientific assistants, useful assistance alone does not imply the ability to independently organize an investigation toward a specified scientific goal \citep{zhang2025scientificmethod}. Scientific research requires more than producing plausible results. 
It requires organizing an investigation that connects a research question to a well-supported scientific conclusion. Doing so involves deciding what evidence to generate and how to interpret and validate it \citep{bogen1988saving,mayo1996error}.
We use \emph{scientific closure} to describe whether such an investigation has adequately resolved the research question, which raises a central question:

\textit{Can agents independently organize an investigation that achieves scientific closure on a specified research question?}

Existing benchmarks evaluate important components of scientific capability, including final answers \citep{rein2024gpqa,phan2025humanity}, executable code and ML-engineering artifacts \citep{tian2024scicode,chan2025mlebench}, and research replication or open-ended research-engineering tasks \citep{starace2025paperbench,wijk2025rebench}. However, they rarely assess whether an investigation as a whole establishes a well-supported answer to the specified research question. An agent may therefore obtain a correct result or reproduce a reported finding without completing the broader investigation, a distinction that existing task-level metrics seldom capture.

\suppressfloats[t]
\begin{table}[t]
\centering
\caption{
Comparison of representative scientific AI benchmarks.
A checkmark indicates direct assessment of a dimension, while ``Partial'' indicates limited or indirect assessment and a cross indicates no explicit assessment.
Scientific closure concerns whether an investigation establishes a well-supported answer to its research question, including the required validation and scope conditions.
}
\label{tab:benchmark_comparison}

\small
\setlength{\tabcolsep}{1.6pt}

\begin{tabular}{
@{}
>{\raggedright\arraybackslash}p{1.30in}
>{\raggedright\arraybackslash}p{1.62in}
*{4}{>{\centering\arraybackslash}p{0.57in}}
@{}
}
\toprule
Benchmark
& Task formulation
& \shortstack{Execution /\\ artifacts}
& \shortstack{Research\\ decisions}
& \shortstack{Evidence\\ integration}
& \shortstack{Scientific\\ closure}
\\
\midrule

GPQA / HLE
& Expert-level QA
& \xmark
& \xmark
& \xmark
& \xmark
\\

CritPt
& Physics reasoning
& \checkmark
& \xmark
& \xmark
& \xmark
\\

SciCode
& Scientific coding
& \checkmark
& \xmark
& \xmark
& \xmark
\\

PaperBench
& Research reproduction
& \checkmark
& Partial
& Partial
& \xmark
\\

RE-Bench
& Research-engineering tasks
& \checkmark
& Partial
& Partial
& \xmark
\\

FrontierScience-R
& Scientific research subproblems
& \xmark
& \checkmark
& Partial
& Partial
\\

PRL-Bench
& Physics research subtasks
& \checkmark
& \checkmark
& \checkmark
& Partial
\\

PhysClosureBench
& Goal-specified physics research
& \checkmark
& \checkmark
& \checkmark
& \checkmark
\\

\bottomrule
\end{tabular}
\end{table}

To address this gap, we formulate goal-specified, agent-directed research tasks in physics, where scientific conclusions are tightly tied to observable evidence, competing physical mechanisms, and well-defined validity regimes. Each task specifies a scientific objective and a bounded setting, while leaving the agent to organize the investigation and determine what evidence is needed to solve the problem. We evaluate agents at two complementary levels that distinguish local scientific accomplishment from overall Research Closure. The \emph{Atomic Score} measures local accomplishment on 20 predefined, task-specific criteria, each worth 0.5 points, that cover formulation, output, identification, mechanism, and validation/scope, while the \emph{Research Closure} (RC0--RC4) evaluates whether the investigation has established the evidence, interpretation, and scope needed for a complete answer.

We instantiate this task formulation and evaluation protocol in \textbf{\textit{PhysClosureBench}}, a benchmark of 50 expert-verified tasks centered on attosecond physics and neighboring regimes, including solid-state strong-field physics, relativistic quantum dynamics, and quantum optics \citep{krausz2009attosecond,kruchinin2018periodic,kockum2019ultrastrong,cruzrodriguez2024quantum}. Evaluating seven frontier models on PhysClosureBench, we characterize their scientific accomplishment and Research Closure. We further examine whether complete closure is achieved across independent runs of the same task. Claude-Opus-5 achieves the highest mean Atomic Score (\MeanScoreOpus/10) and RC4 rate (\RCFourPercentOpus\%), whereas Gemini-3.1-Pro obtains the lowest (2.85/10 and 0.7\%, respectively). Even for the strongest model, complete closure remains uncommon. Achieving complete closure reliably can be even harder. For Claude-Opus-5, only \AllRCFourOpus\% of tasks reach RC4 in all three independent runs. These results reveal a substantial gap between local accomplishment and complete closure, and a further gap between demonstrating the ability to achieve closure and doing so reliably across independent runs.

Our contributions are threefold:
\begin{enumerate}
    \interlinepenalty=10000
    \setlength{\topsep}{3pt}
    \setlength{\itemsep}{1pt}
    \setlength{\parsep}{0pt}
    \setlength{\parskip}{0pt}

    \item We introduce PhysClosureBench, a benchmark of goal-specified physics research tasks that evaluates whether agents can carry investigations through to well-supported answers to specified research questions.

    \item We develop an evaluation framework with two complementary perspectives. An Atomic Score measures local scientific accomplishment across five functional dimensions, while Research Closure assesses how fully an investigation satisfies its task-specific scientific requirements.

    \item We evaluate seven frontier models and find that substantial local accomplishment does not ensure complete closure, nor does achieving closure in one run guarantee the same outcome in other independent runs.

\end{enumerate}


\FloatBarrier
\section{Related Work}

Recent scientific AI benchmarks span expert-level reasoning and increasingly complex research activities. GPQA and Humanity's Last Exam assess expert-level question answering \citep{rein2024gpqa,phan2025humanity}. Moving toward research-level scientific problems, CritPt evaluates composite frontier-physics challenges and finer-grained checkpoints with machine-verifiable answers, including code-enabled evaluation \citep{zhu2025critpt}. FrontierScience's Research track assesses expert-authored, PhD-level scientific subproblems using granular rubrics \citep{wang2026frontierscience}. Beyond scientific reasoning, agents such as The AI Scientist and The AI Scientist-v2 explore broader research workflows involving experiment design, execution, and iterative analysis \citep{lu2024aiscientist,yamada2025aiscientistv2}. Related benchmarks evaluate different aspects of executable research. SciCode focuses on scientific code generation \citep{tian2024scicode}; ScienceAgentBench assesses generated programs and execution results in data-driven scientific tasks \citep{chen2025scienceagentbench}; MLE-Bench and RE-Bench evaluate machine-learning and research-engineering tasks, while PaperBench examines research reproduction \citep{chan2025mlebench,wijk2025rebench,starace2025paperbench}. More closely related to physics research, PRL-Bench evaluates exploration-oriented, long-horizon theoretical and computational tasks with code execution and rubric-based assessment \citep{miao2026prlbench}. Table~\ref{tab:benchmark_comparison} summarizes representative differences in their evaluation scope and design.

PhysClosureBench instead makes scientific closure at the level of a goal-specified, agent-directed investigation an explicit evaluation target. It distinguishes local scientific accomplishment from whether the evidence, interpretation, and validation established by the agent are sufficient to resolve the original research question, and examines how consistently complete closure is achieved across independent attempts.

\section{Benchmark and Evaluation Design}

\begin{figure}[!t]
\centering
\includegraphics[width=\linewidth]{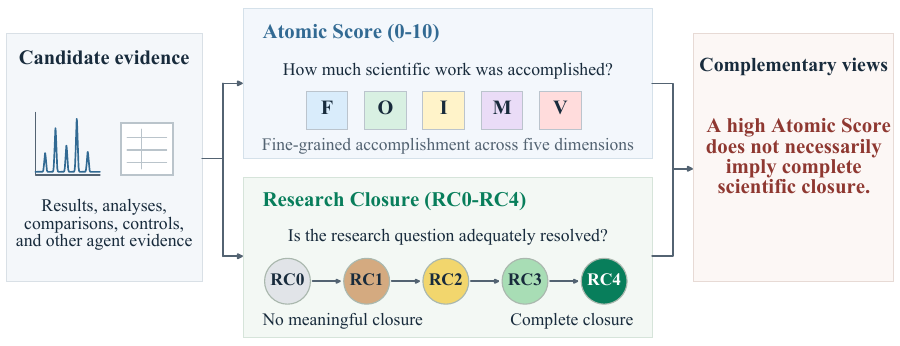}
\caption{\textbf{Two complementary views of scientific performance.} Atomic Score (0--10) measures local accomplishment across F/O/I/M/V. Research Closure (RC0--RC4) assesses how fully the investigation resolves the research question. A high Atomic Score does not guarantee complete closure.}
\label{fig:performance_views}
\par\vspace{8pt}
\begingroup
\begin{minipage}[t]{0.42\linewidth}
\centering
\captionof{table}{Atomic scoring dimensions.}
\label{tab:atomic_dimensions}
\footnotesize
\setlength{\tabcolsep}{2.8pt}
\setlength{\fboxsep}{1.2pt}
\renewcommand{\arraystretch}{1.12}

\begin{tabularx}{\linewidth}{
@{}
>{\raggedright\arraybackslash}p{0.43\linewidth}
>{\raggedright\arraybackslash}X
@{}
}
\toprule
Dimension & Scientific question \\
\midrule

\colorbox[HTML]{D9ECFB}{\strut\textbf{F} Formulation}
& Physical and numerical setup correct? \\

\colorbox[HTML]{D8F0E2}{\strut\textbf{O} Output}
& Required observables generated correctly? \\

\colorbox[HTML]{FFF3C9}{\strut\textbf{I} Identification}
& Relevant features identified correctly? \\

\colorbox[HTML]{EADCF7}{\strut\textbf{M} Mechanism}
& Mechanism supported by evidence? \\

\colorbox[HTML]{FFDCDC}{\strut\textbf{V} Validation/Scope}
& Applicable controls and scope justified? \\

\bottomrule
\end{tabularx}
\end{minipage}\hfill
\begin{minipage}[t]{0.56\linewidth}
\centering
\captionof{table}{Research Closure levels.}
\label{tab:rc_levels}
\footnotesize
\setlength{\tabcolsep}{2.5pt}
\setlength{\fboxsep}{1.2pt}
\renewcommand{\arraystretch}{1.12}

\begin{tabularx}{\linewidth}{
@{}
l
>{\raggedright\arraybackslash}X
@{}
}
\toprule
Level & Interpretation \\
\midrule

\colorbox{green!20}{\strut\textbf{RC4}}
& \textbf{Complete closure.} Required evidence and scope conditions satisfied. \\

\colorbox{green!10}{\strut\textbf{RC3}}
& \textbf{Main conclusion supported.} Material validation or scope gaps remain. \\

\colorbox{yellow!20}{\strut\textbf{RC2}}
& \textbf{Partial investigation.} Core evidence present; central mechanism or other key requirement unresolved. \\

\colorbox{orange!18}{\strut\textbf{RC1}}
& \textbf{Limited investigation.} Essential primary output or branch missing. \\

\colorbox{gray!15}{\strut\textbf{RC0}}
& \textbf{No meaningful closure.} Central scientific object invalid or absent. \\

\bottomrule
\end{tabularx}
\end{minipage}
\endgroup
\end{figure}
\def\MainEvaluationTables{}

\subsection{Task Design and Agent-Directed Setting}

PhysClosureBench contains 50 goal-specified physics research tasks derived from published studies in authoritative journals in physics such as \textit{Physical Review Letters} and \textit{Physical Review A}, spanning strong-field atomic and molecular physics, attosecond science, solid-state and condensed-matter physics, quantum optics, and neighboring quantum light--matter regimes.

Each task specifies the scientific objective and the information necessary to make the problem well-posed, including its physical setting, relevant parameters, constraints, and computational resources. The benchmark fixes what scientific question must be addressed, while leaving the overall investigation strategy open within the task-specific scientific constraints. No complete solution workflow or fixed diagnostic and validation sequence is prescribed. Agents instead determine how to carry out the investigation, including the numerical implementation, observables, comparisons, controls, and validation procedures needed to answer the question.

Tasks are executed in isolated, network-disabled computational environments under task-specific CPU and wall-time limits. Within these resource limits, agents decide how computational effort is allocated across planning, exploration, analysis, and validation.

\subsection{Task Reliability and Verification}

We use a multi-stage verification process to ensure that the benchmark tasks are scientifically grounded, executable, and appropriately resourced. First, as mentioned before, every task is derived from a published physics study, providing a traceable basis for the target phenomenon, physical regime, and scientific conclusion. Second, a domain expert designed and supervised the semi-automated pipeline used for task construction and rubric design. Review by other experts checked that the intended scientific questions and evidential requirements were preserved.

Third, the source results underlying every task were manually reproduced before inclusion in the benchmark. We verified that the relevant numerical results, physical trends, and scientific conclusions could be recovered under the specified conditions. These reproductions were also used to calibrate the resource limits. CPU allocations and wall-time limits were determined from the observed reproduction costs, with additional allowance for planning, exploratory calculations, method selection, and unsuccessful intermediate attempts. This process ensures that tasks are feasible under the provided resources without constraining agents to the reference reproduction pathway.

\subsection{Evaluation Framework}

Scientific research performance can be assessed at two complementary levels. One is the extent of scientific work accomplished during the research process, and the other is whether the investigation as a whole reaches a well-supported answer to the research question. Substantial progress on individual research components does not necessarily imply complete closure, and closure may be achieved through different valid research pathways as long as the task-specific scientific requirements are satisfied. Figure~\ref{fig:performance_views} summarizes these two complementary perspectives.


\begin{enumerate}
    \item \textit{\textbf{Atomic scientific accomplishment.}} The Atomic Score measures local accomplishment across 20 task-specific, predefined half-point criteria, organized into five functional dimensions, including Formulation (F), Output (O), Identification (I), Mechanism (M), and Validation/Scope (V), as summarized in Table~\ref{tab:atomic_dimensions}. The resulting 0--10 score captures the extent of relevant scientific work completed, without requiring that the investigation itself be complete.
    
    \item \textit{\textbf{Research Closure.}} Research Closure (RC0--RC4) evaluates the overall completion of the investigation relative to the research question. The closure levels, summarized in Table~\ref{tab:rc_levels}, are determined by task-specific scientific requirements rather than by a fixed workflow or threshold on the Atomic Score. The distinction between adjacent RC levels depends on the role of an unresolved requirement in supporting the main scientific conclusion, including its task-specific prerequisites, rather than on its assignment to a particular functional dimension.
\end{enumerate}

Atomic Scores provide fine-grained information about where scientific progress is made, whereas Research Closure assesses whether the investigation meets the requirements for a complete scientific answer. Together, they support both diagnostic analysis of research behavior and consistent comparison across models under a common evaluation protocol.

\begin{figure}[!t]
    \centering
    \includegraphics[width=\linewidth]{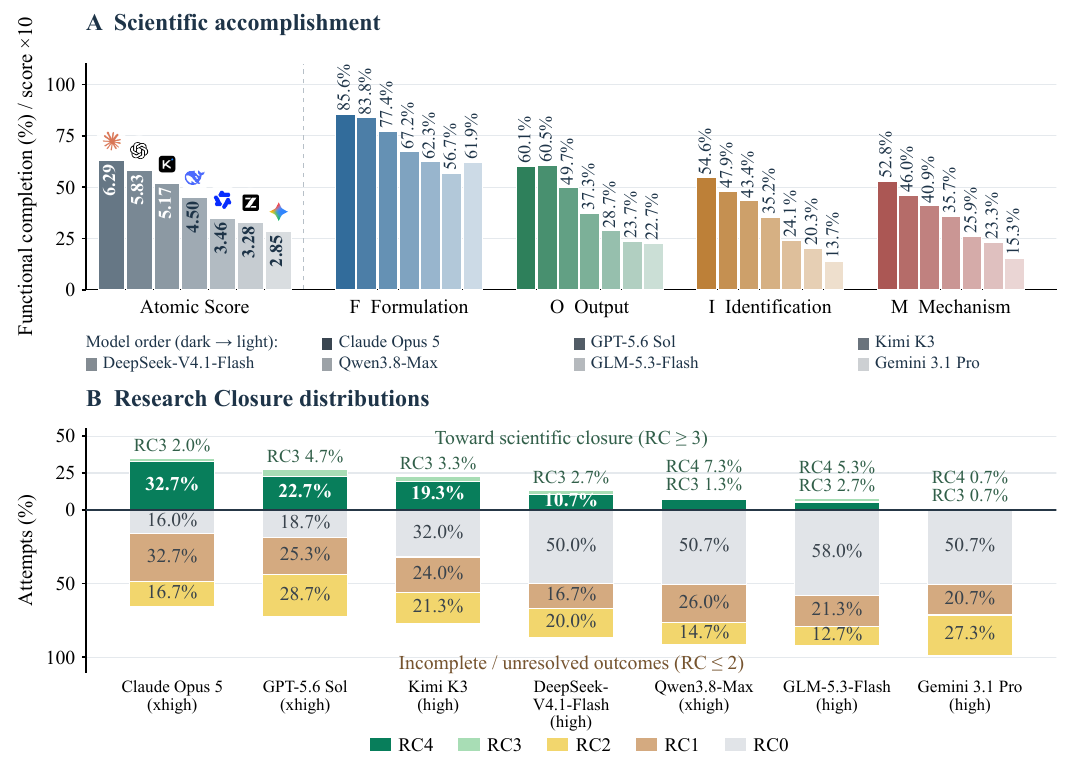}
    \caption{
    \textbf{Scientific accomplishment and Research Closure across frontier models.}
    \textbf{(A)} Mean Atomic Score and functional completion for Formulation (F), Output (O), Identification (I), and Mechanism (M). Atomic Scores are shown on the shared axis after multiplication by 10; functional completion is the mean attempt-level fraction of applicable criteria satisfied within each dimension.
    \textbf{(B)} Distribution of Research Closure levels across independent attempts. RC4 denotes complete closure, RC3 indicates that the main conclusion is supported, but material validation or scope gaps remain, and RC0--RC2 indicate different degrees of incomplete investigation.
    }
    \label{fig:scientific_accomplishment}
\end{figure}

\Needspace{7\baselineskip}
\section{Experimental Setup}

\subsection{Models and Evaluation Procedure}

We evaluate seven frontier models---Claude-Opus-5~\citep{anthropic2026claudeopus5}, GPT-5.6-Sol~\citep{openai2026gpt56sol}, Kimi-K3~\citep{kimiteam2026kimik3}, DeepSeek-V4.1-Flash~\citep{deepseekai2026v41flash}, Qwen-3.8-Max~\citep{alibabagroup2026qwen38max}, GLM-5.3-Flash~\citep{zai2026glm53flash}, and Gemini-3.1-Pro~\citep{googledeepmind2026gemini31pro}---on all 50 tasks, with three independent runs for each model--task pair, yielding 1,050 final scored attempts. The evaluated configurations use xhigh reasoning effort for Claude-Opus-5, GPT-5.6-Sol, and Qwen-3.8-Max, and high reasoning effort for Kimi-K3, DeepSeek-V4.1-Flash, GLM-5.3-Flash, and Gemini-3.1-Pro.

Each attempt runs in a separate OpenCode~\citep{anomaly2026opencode}
session with its own workspace, initialized with the task specification, resource limits, and common instructions on tool use, research conduct,
and submission requirements. Agents can read and modify files in their workspaces and execute scientific code through a custom Bash interface in an isolated, network-disabled container. When a run ends or reaches its wall-time limit, the controller collects the report and supporting artifacts available at that point for grading, including assessable partial results. Appendix~\ref{app:benchmark_tasks} details the environment and protocol.

Incomplete work produced by the agent is included in the evaluation. Unsuccessful calculations, abandoned approaches, and incomplete analyses are scored according to the evidence produced by the agent. Runs that reach the time limit are graded on the scientific work available at that point, with no additional penalty for timeout itself. When an attempt produces no assessable scientific result, it receives zero atomic credit and RC0. Executions invalidated by infrastructure failures are excluded from scoring and rerun to obtain a valid measurement.

All assessable submissions are evaluated using a staged automated protocol with GPT-5.6-Sol(high). Each submission first receives an independent first-pass grade using the fixed evaluation prompt and task-specific rubric, followed by a targeted review with access to the initial assessment. Conditional adjudication is triggered only when a residual material disagreement could affect the Atomic Score, Research Closure, or attempt validity. Evaluation is restricted to evidence produced by the agent. Submitted artifacts and execution records may be inspected for verification, but the evaluator's checks cannot substitute for analyses or interpretations missing from the submissions.

Five physics-trained experts reviewed 100 evaluation cards covering all 50 tasks, yielding 200 selected criterion-level judgments. Final human assessments agreed with automated grading in 81.8\% of the selected atomic-criterion judgments, with 75.9\% agreement on criteria explicitly linked to Research Closure. 
Appendix~\ref{app:grading_protocol} describes the review protocol and the closure-relevant criteria in detail.

\subsection{Analysis Measures}

The individual research attempt is the basic unit of analysis. Unless otherwise specified, model-level results are aggregated over three independent runs of each of the 50 benchmark cases (150 attempts per model). We report the mean Atomic Score and the distribution of Research Closure levels, including the overall RC4 rate, as the primary measures of scientific accomplishment and closure.

For Formulation, Output, Identification, and Mechanism, functional completion is computed within each attempt as the fraction of satisfied criteria among those applicable to that attempt. Model-level completion is the mean of the attempt-level completion rates, giving equal weight to each research attempt. Attempts with no assessable scientific accomplishment contribute zero completion. Validation/Scope is analyzed separately because its applicability depends more directly on the method used, evidence produced, and resulting conclusion of each attempt.

To characterize consistency of Research Closure across independent runs, we additionally report, for each model, the fraction of the 50 tasks that achieve RC4 in at least one of the three independent runs and the fraction that achieve RC4 in all three runs. Appendix~\ref{app:statistical_analysis} provides the detailed
calculations for the analysis measures, including
the treatment of N/A criteria and how results are aggregated.

\FloatBarrier
\begin{figure}[!t]
    \centering
    \includegraphics[width=\linewidth]{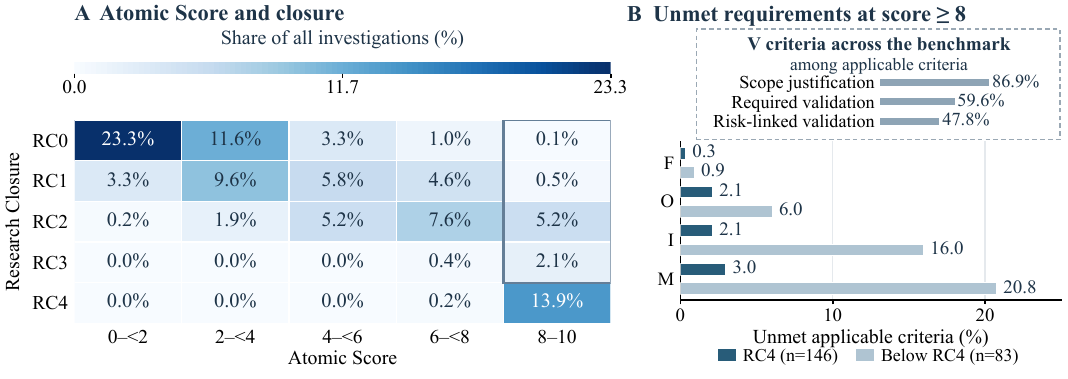}
    \caption{
    \textbf{Relationship between local scientific accomplishment and Research Closure.}
    \textbf{(A)} Joint distribution of Atomic Score and Research Closure across all 1,050 independent attempts. Each cell reports the percentage of all attempts falling into the corresponding score bin and RC level. The highlighted region marks high-scoring attempts below RC4.
    \textbf{(B)} Unmet applicable criteria for Formulation (F), Output (O), Identification (I), and Mechanism (M) among the 229 attempts scoring at least 8/10, comparing RC4 ($n=146$) with below RC4 ($n=83$). Bars show 100\% minus the mean attempt-level functional completion within each group. The inset reports the fraction of satisfied applicable Validation/Scope (V) criteria across the full benchmark, rather than within the high-scoring groups; N/A criteria are excluded.
    }
    \label{fig:closure_association}
\end{figure}

\section{Results and Discussion}

We analyze scientific-agent performance in terms of local scientific accomplishment, Research Closure, the evidence patterns underlying closure, and consistency of Research Closure across independent runs.

\subsection{Scientific Accomplishment and Research Closure}

We first compare local scientific accomplishment with Research Closure across models, as shown in Figure~\ref{fig:scientific_accomplishment}. Claude-Opus-5 achieves the highest mean Atomic Score at \MeanScoreOpus/10, followed by GPT-5.6-Sol at \MeanScoreGPT/10 and Kimi-K3 at 5.17/10, while the remaining models score 4.50 or below. The functional profiles further reveal distinct patterns across research capabilities. Formulation has the highest completion rate among F/O/I/M for every model, suggesting that current systems are comparatively reliable at establishing an appropriate scientific setup. For Claude-Opus-5, completion reaches 85.6\% in Formulation, compared with 60.1\% in Output, 54.6\% in Identification, and 52.8\% in Mechanism. The relative advantage of stronger models is particularly visible in Output. Claude-Opus-5, GPT-5.6-Sol, and Kimi-K3 achieve 60.1\%, 60.5\%, and 49.7\%, respectively, compared with 37.3\% or less for the remaining models. In contrast, Identification and Mechanism remain challenging across the model set, with even the strongest model completing only about half of the applicable criteria in either dimension. Some submissions lack reliable upstream results needed for interpretation; others generate relevant evidence but fail to identify its key features or establish a supported physical explanation.

Research Closure broadly follows the model ranking by Atomic Score, but complete closure remains uncommon even among the leading models. Claude-Opus-5, GPT-5.6-Sol, and Kimi-K3 achieve RC4 in \RCFourPercentOpus\%, 22.7\%, and 19.3\% of attempts, respectively. Despite their stronger performance on individual scientific requirements, most of their investigations do not reach complete closure. The full RC distribution also reveals differences that RC4 rates alone cannot capture. DeepSeek-V4.1-Flash reaches RC4 in 10.7\% of attempts, more frequently than the three lower-scoring models, yet 50.0\% of its attempts remain at RC0. Thus, a comparatively higher complete-closure rate can coexist with a large proportion of investigations that fail to establish the essential scientific basis of the task.

Taken together, the two panels highlight \textbf{a substantial gap between local scientific accomplishment and complete closure}.

\subsection{High Atomic Scores and Incomplete Research Closure}

We next examine how progress on individual scientific requirements relates to and differs from the overall completion of an investigation (Figure~\ref{fig:closure_association}).

Panel~A shows that higher Atomic Scores generally correspond to higher Research Closure levels. This overall agreement is expected, as both measures assess how well an investigation meets its scientific requirements. Yet a high Atomic Score alone cannot establish whether the research question has been resolved. Among 229 attempts scoring at least 8/10, 83 (36.2\%) do not reach RC4. Of these, 55 remain at RC2, indicating that substantial local progress can coexist with unresolved requirements for the main scientific conclusion, rather than merely incomplete validation.

Panel~B provides a closer look at these remaining gaps. High-scoring attempts below RC4 satisfy nearly all Formulation and Output criteria, but leave 16.0\% of Identification and 20.8\% of Mechanism criteria unmet. The corresponding fractions among RC4 attempts are only 2.1\% and 3.0\%. This suggests that producing the required results does not necessarily mean that their scientific implications are established. For example, one investigation generated the requested spectra and separated their contributions to the signal, but failed to connect these results to the underlying physical mechanism. Most of the numerical work was complete, yet a key part of the explanation remained missing.

A different limitation arises when the main mechanism is supported, but the validity of the conclusion remains uncertain. The Validation/Scope (V) results, shown separately in the inset of Panel~B, capture this aspect of scientific closure. Across the benchmark, scope justification is satisfied more frequently (86.9\%) than required validation (59.6\%) or risk-linked validation (47.8\%). In another high-scoring investigation, the main mechanism was established, but an applicability concern identified through the agent's own calculations remained unresolved. The remaining uncertainty concerned the validity of the conclusion, rather than the mechanism itself. Addressing such concerns requires appropriate validation or restricting the conclusion to a regime supported by the evidence.

\begin{figure}[!t]
    \centering
    \includegraphics[width=\linewidth]{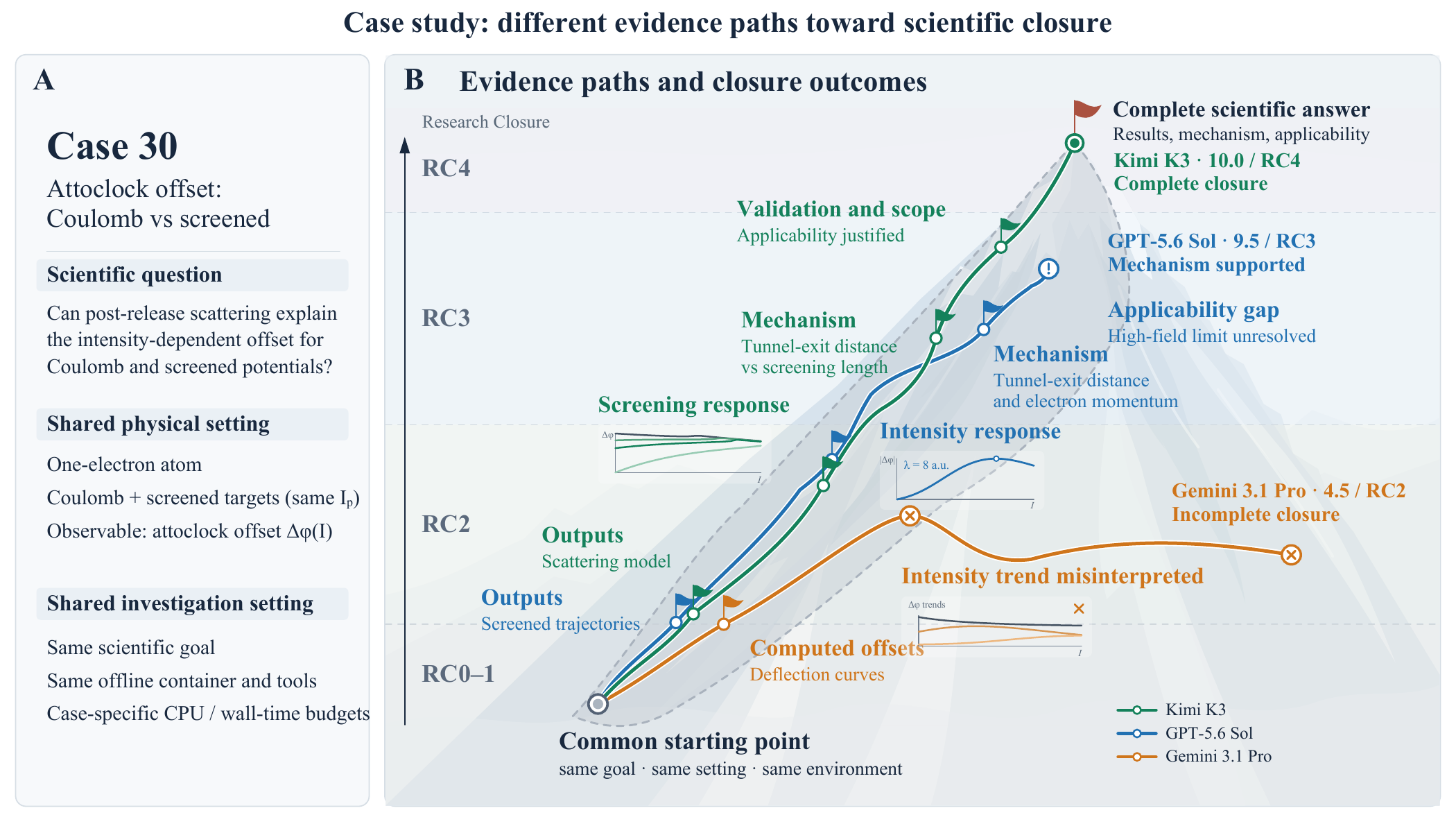}
    \caption{
    \textbf{{Representative evidence and scientific closure.}}
    Case~30 compares representative runs from Kimi-K3, GPT-5.6-Sol, and Gemini-3.1-Pro, which reached RC4, RC3, and RC2, respectively. The shaded region illustrates a possible progression from numerical evidence to a supported conclusion; Gemini's deviation represents the misinterpretation of the computed intensity trends. Paths are schematic, not execution times, Atomic Scores, or intermediate RC assignments. Inset plots schematically illustrate trends in the submitted results. Appendix~\ref{app:representative_trajectories} documents the research activities and evidence.
    }
    \label{fig:representative_pathways}
\end{figure}

\Needspace{10\baselineskip}
\subsection{Evidence Pathways toward Scientific Closure}

Figure~\ref{fig:representative_pathways} compares three investigations of the same attoclock problem. How does post-release scattering change an electron's angular offset as laser intensity and screening length vary? {\citep{pcbSourceCase30}} All three calculate Coulomb and screened potentials matched to the same binding energy. The task requires interpreting the screening-dependent intensity response and establishing how far the reduced scattering model supports that interpretation.

Gemini-3.1-Pro generates relevant outputs but describes increasing or turning-over screened responses as monotonically decreasing, leaving the main trend and its mechanism unsupported (RC2). GPT-5.6-Sol identifies and explains the response changes, but does not resolve the high-field applicability risk exposed by its own Coulomb barrier diagnostic (RC3). Kimi-K3 corrects an angle-sign inconsistency, explains the competition between exit radius and screening length, and justifies its applicability limits (RC4). Appendix~\ref{app:representative_trajectories} documents these scientific revisions, numerical results, and applicability limits.

The comparison distinguishes an incorrect interpretation of computed trends from a supported mechanism with unresolved applicability limits.

\Needspace{0.43\textheight}
\subsection{{Research Closure across Repeated Runs}}
\noindent
\begin{minipage}[t]{0.53\linewidth}
\vspace{0pt}
\setlength{\parskip}{0.25\baselineskip}
A model may reach RC4 on a task in one run but fall short in another. Figure~\ref{fig:closure_reproducibility} compares the proportion of tasks reaching RC4 at least once with the proportion reaching RC4 in all three independent runs.

    Claude-Opus-5 reaches RC4 at least once on 50.0\% of tasks, but on only 18.0\% in all three runs. GPT-5.6-Sol and Kimi-K3 each reach RC4 at least once on 40.0\% of tasks, but reach RC4 in all three runs on only 10.0\% and 4.0\% of tasks, respectively. These two measures distinguish a model's demonstrated ability to reach closure on a task from the reliability of achieving closure across independent runs.

    DeepSeek-V4.1-Flash reaches RC4 at least once on 24.0\% of tasks, but on none in all three runs. Requiring RC4 across all three independent investigations is substantially more stringent than achieving it at least once. Even the strongest models can demonstrate that closure is achievable on a task without achieving it reliably across repeated investigations.
\end{minipage}\hfill
\begin{minipage}[t]{0.45\linewidth}
\vspace{0pt}
\centering
\captionsetup{type=figure,font=small,skip=2pt}
\includegraphics[width=\linewidth]{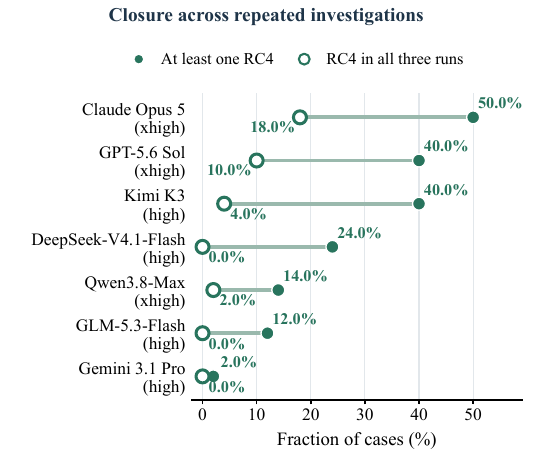}
\caption{\textbf{Research Closure across repeated runs.}
    For each model, filled markers show the fraction of the 50 tasks reaching RC4 in at least one of three independent runs, while open markers show the fraction reaching RC4 in all three runs.}
\label{fig:closure_reproducibility}
\end{minipage}
\par

\section{Conclusion}
\label{sec:conclusion}

We introduced PhysClosureBench, a benchmark for evaluating scientific-agent performance beyond answer correctness or artifact generation. The benchmark combines goal-specified physics research tasks with an evaluation framework that assesses both atomic scientific accomplishment and Research Closure. Across seven frontier models, we found that higher Atomic Scores generally accompanied higher closure levels, yet high-scoring investigations did not always achieve complete closure. Case-level analysis further showed that investigations can fail for qualitatively different reasons, including misinterpreted results, unsupported physical mechanisms, and unresolved applicability limits. Repeated runs revealed a further gap between demonstrating the ability to achieve closure and achieving it reliably.
These findings highlight the need to evaluate not only how much scientific work agents accomplish, but also whether their investigations progress from a research question to a well-supported conclusion, and how reliably they achieve complete closure across independent runs.

\section*{Acknowledgments}

This work was supported by the Shanghai Artificial Intelligence Laboratory. We are grateful to the authors and open-source communities whose work made this project possible.


\FloatBarrier
\setlength{\bibsep}{1pt}
\bibliography{references}
\bibliographystyle{iclr2026_conference}

\appendix

\section{Benchmark Tasks}
\label{app:benchmark_tasks}

\begin{figure}[bp]
    \centering
    \includegraphics[width=\linewidth]{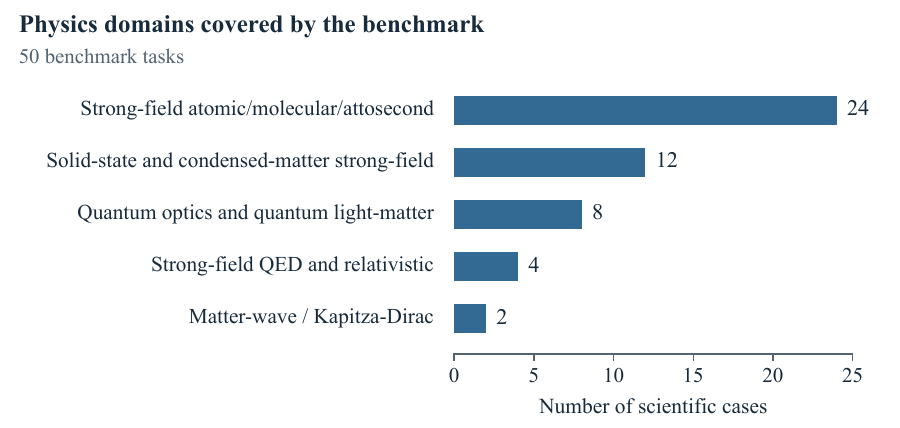}
    \caption{Domain distribution of the 50 benchmark tasks.}
    \label{fig:appendix_taxonomy}
\end{figure}

\subsection{Benchmark Construction and Quality Control}

PhysClosureBench uses a semi-automated construction process with expert guidance. A domain expert with research experience in ultrafast and strong-field physics designed, iteratively calibrated, and supervised the task-construction pipeline. Objectives based on published studies were formulated into bounded research tasks. Structured model-assisted generation and multi-pass checks supported task formulation, rubric construction, and consistency checks against the source paper.

During pipeline development and benchmark construction, review by other domain experts assessed task validity, grading criteria, scientific requirements and their prerequisites, acceptable alternative evidence, and the separation between task instructions given to agents and hidden grading criteria. Additional domain experts provided scientific feedback and checks on representative tasks and benchmark design choices. Successful reference reproduction was required before a task was admitted to the benchmark, confirming that the intended physical behavior could be reproduced within the benchmark setting.

For each case, the task specification, resource limits, and grading rubric were finalized before the model evaluation.

\subsection{Task-Specific Resource Limits}

All 50 cases use isolated, network-disabled, single-node environments. Table~\ref{tab:case_resource_limits} summarizes the CPU-core allocations and wall-time limits for all 50 cases. These limits define the computational resources available for each investigation.

\begin{table}[tbp]
\centering
\caption{Case-level CPU and wall-time limits for the 50 benchmark tasks.}
\label{tab:case_resource_limits}
\footnotesize
\setlength{\tabcolsep}{4pt}
\begin{tabular}{@{}lrr@{\hspace{24pt}}lrr@{}}
\toprule
Case & CPU cores & Minutes & Case & CPU cores & Minutes \\
\midrule
Case01 & 8 & 30 & Case26 & 16 & 90 \\
Case02 & 16 & 30 & Case27 & 8 & 60 \\
Case03 & 32 & 240 & Case28 & 16 & 60 \\
Case04 & 8 & 30 & Case29 & 32 & 90 \\
Case05 & 32 & 45 & Case30 & 8 & 45 \\
Case06 & 8 & 45 & Case31 & 96 & 360 \\
Case07 & 8 & 45 & Case32 & 96 & 720 \\
Case08 & 32 & 60 & Case33 & 8 & 45 \\
Case09 & 16 & 30 & Case34 & 96 & 180 \\
Case10 & 8 & 30 & Case35 & 96 & 120 \\
Case11 & 16 & 60 & Case36 & 8 & 45 \\
Case12 & 16 & 60 & Case37 & 8 & 45 \\
Case13 & 8 & 45 & Case38 & 16 & 90 \\
Case14 & 8 & 45 & Case39 & 96 & 120 \\
Case15 & 8 & 60 & Case40 & 16 & 90 \\
Case16 & 8 & 45 & Case41 & 96 & 180 \\
Case17 & 16 & 45 & Case42 & 96 & 180 \\
Case18 & 8 & 45 & Case43 & 32 & 90 \\
Case19 & 16 & 60 & Case44 & 8 & 45 \\
Case20 & 32 & 90 & Case45 & 32 & 90 \\
Case21 & 8 & 45 & Case46 & 16 & 90 \\
Case22 & 8 & 60 & Case47 & 8 & 45 \\
Case23 & 8 & 45 & Case48 & 16 & 60 \\
Case24 & 8 & 60 & Case49 & 8 & 60 \\
Case25 & 8 & 60 & Case50 & 8 & 45 \\
\bottomrule
\end{tabular}
\end{table}

\paragraph{Resource calibration.}
Successful reference reproductions were used to calibrate the CPU and wall-time budgets for all 50 cases. The assigned limits were based on observed reproduction costs, with additional allowance for planning, exploratory calculations, method selection, and unsuccessful intermediate attempts. The resulting resource limits were intended to provide sufficient resources for a scientifically viable investigation while maintaining a bounded execution setting. 

\FloatBarrier

\subsection{Agent Execution Environment and Examination Protocol}
\label{app:execution_protocol}

Each attempt was conducted in a separate OpenCode session and workspace. The OpenCode deployment used version 1.18.17. Slurm allocated task-specific computational resources, while scientific code was executed through a Bash interface in an isolated, network-disabled container. Model inference was handled separately through the configured model service. The environment provided numerical Python libraries and compiled-language tools, allowing agents to implement, execute, and analyze their own calculations.

At the start of each run, agents received common instructions and task-specific research materials. \texttt{EXAM\_INSTRUCTIONS.md} provided the common examination rules and submission requirements. \texttt{CANDIDATE\_ENVIRONMENT.md} described the available computational tools. The \texttt{task/} directory contained the case-specific research specification and resource limits, with accompanying task materials where supplied. These defined the scientific objective, physical setting, parameters, and constraints. The Case~30 specification is reproduced in Appendix~\ref{app:worked_case30}. The retained common initial instruction template was:

\begin{casebox}{Common initial instruction}

You are an isolated research-exam candidate in case \{case\_id\}, attempt \{attempt\}.

Read EXAM\_INSTRUCTIONS.md and every supplied file under task/ before deciding on your research route. Complete the research task independently. You may derive equations, write code, run Linux commands, generate data and figures, and revise your approach within the visible time and resource limits.

All final answer materials must be placed under submission/ using the required entry-point filenames. Keep scratch work under scratch/. Do not ask the user or examiner for guidance. Do not use network search, external agents, hidden files, previous submissions, target-paper materials, or unrecorded references. If the calculation cannot be completed, submit the strongest auditable partial result and document the blocker rather than fabricating data.

Before finishing, verify that submission/ contains the required reports, code/data/figure artifacts needed to audit your claims, and honest reference/integrity declarations.
\end{casebox}

Candidates could query the remaining time and choose their own computational, analytical, and validation procedures within these requirements.

Standard submission files documented the scientific report, derivations, execution history, references, integrity declarations, and links between reported claims and supporting artifacts (Appendix~\ref{app:worked_case30}). Agents organized task-specific code, data, and figures. At completion or the work deadline, the controller collected the available submission for evaluation without allowing further scientific work during collection.

\FloatBarrier

\section{Grading Protocol}
\label{app:grading_protocol}

\ifdefined\MainEvaluationTables\else
\begin{table}[H]
\centering
\begin{minipage}[t]{0.42\linewidth}
\centering
\captionof{table}{Atomic scoring dimensions.}
\label{tab:atomic_dimensions}
\footnotesize
\setlength{\tabcolsep}{2.8pt}
\setlength{\fboxsep}{1.2pt}
\renewcommand{\arraystretch}{1.12}

\begin{tabularx}{\linewidth}{
@{}
>{\raggedright\arraybackslash}p{0.43\linewidth}
>{\raggedright\arraybackslash}X
@{}
}
\toprule
Dimension & Scientific question \\
\midrule

\colorbox[HTML]{D9ECFB}{\strut\textbf{F} Formulation}
& Physical and numerical setup correct? \\

\colorbox[HTML]{D8F0E2}{\strut\textbf{O} Output}
& Required observables generated correctly? \\

\colorbox[HTML]{FFF3C9}{\strut\textbf{I} Identification}
& Relevant features identified correctly? \\

\colorbox[HTML]{EADCF7}{\strut\textbf{M} Mechanism}
& Mechanism supported by evidence? \\

\colorbox[HTML]{FFDCDC}{\strut\textbf{V} Validation/Scope}
& Applicable controls and scope justified? \\

\bottomrule
\end{tabularx}
\end{minipage}\hfill
\begin{minipage}[t]{0.56\linewidth}
\centering
\captionof{table}{Research Closure levels.}
\label{tab:rc_levels}
\footnotesize
\setlength{\tabcolsep}{2.5pt}
\setlength{\fboxsep}{1.2pt}
\renewcommand{\arraystretch}{1.12}

\begin{tabularx}{\linewidth}{
@{}
l
>{\raggedright\arraybackslash}X
@{}
}
\toprule
Level & Interpretation \\
\midrule

\colorbox{green!20}{\strut\textbf{RC4}}
& \textbf{Complete closure.} Required evidence and scope conditions satisfied. \\

\colorbox{green!10}{\strut\textbf{RC3}}
& \textbf{Main conclusion supported.} Material validation or scope gaps remain. \\

\colorbox{yellow!20}{\strut\textbf{RC2}}
& \textbf{Partial investigation.} Core evidence present; central mechanism or other key requirement unresolved. \\

\colorbox{orange!18}{\strut\textbf{RC1}}
& \textbf{Limited investigation.} Essential primary output or branch missing. \\

\colorbox{gray!15}{\strut\textbf{RC0}}
& \textbf{No meaningful closure.} Central scientific object invalid or absent. \\

\bottomrule
\end{tabularx}
\end{minipage}
\end{table}
\fi

\subsection{Automated Grading and Scoring Semantics}

Tables~\ref{tab:atomic_dimensions} and~\ref{tab:rc_levels} summarize the functional dimensions and Research Closure states used throughout the evaluation. Each case has a grading rubric specifying twenty binary half-point atoms together with the task-specific scientific requirements and their prerequisite structure. Conditional criteria are activated according to the method, evidence, or conclusion produced by the candidate. 

GPT-5.6-Sol at high reasoning effort serves as the automated evaluator throughout the staged grading protocol. Each assessable submission first receives an independent first-pass grade using the fixed evaluation prompt and case-specific rubric. A targeted review then checks grading decisions affecting the score or RC assessment, the effects of unresolved prerequisites, total score, Research Closure, and attempt validity with access to the first-pass assessment. Conditional adjudication is triggered only when a residual material disagreement could affect the final score, Research Closure, or validity of the attempt. Appendix~\ref{app:grading_instructions} provides the common grading rules, first-pass and targeted-review prompts, and the conditional adjudication procedure.

The Atomic Score is the sum of twenty fixed half-point atoms.
Some criteria in the case-specific rubric apply only under
specified circumstances. If these circumstances do not arise in a
scientifically assessable attempt, the corresponding criteria are marked N/A.
An N/A criterion retains its fixed 0.5-point
contribution to the 10-point score.
This convention avoids penalizing an attempt for an inapplicable requirement; it does not imply that the corresponding scientific work was performed. Therefore, the N/A criterion is excluded from both the numerator and denominator of applicable-criterion functional completion. 
For risk-triggered Validation/Scope criteria, N/A therefore indicates only that the triggering risk did not arise, not that positive validation evidence was produced. 
A run with no scientifically assessable submission (a terminal-zero outcome) receives score 0 and RC0 directly, before atom-level scoring, and receives no N/A credit.

Table~\ref{tab:atom_allocation} summarizes the allocation of all 1,000 rubric atoms across the five functional dimensions and their variation across the 50 tasks. Each atom is worth 0.5 points, so the number assigned to a dimension determines its share of each task's 10-point maximum.

\begin{table}[tbp]
\centering
\caption{Atom allocation across 50 cases. Total atom counts across all cases; Mean and Median are calculated from the 50 per-case atom counts for each dimension; Range gives their minimum and maximum. Mean score weight is the average percentage of each task's 10-point maximum allocated to a dimension.}
\label{tab:atom_allocation}
\footnotesize
\setlength{\tabcolsep}{3pt}
\begin{tabular}{@{}lrrrrr@{}}
\toprule
Dimension & Total & Mean & Median & Range & Mean score weight \\
\midrule
F & 164 & 3.28 & 3 & 2--4 & 16.4\% \\
O & 192 & 3.84 & 4 & 2--6 & 19.2\% \\
I & 163 & 3.26 & 3 & 3--4 & 16.3\% \\
M & 369 & 7.38 & 7 & 7--9 & 36.9\% \\
V & 112 & 2.24 & 2 & 1--3 & 11.2\% \\
\midrule
Total & 1000 & 20.00 & 20 & 20--20 & 100.0\% \\
\bottomrule
\end{tabular}
\end{table}

Research Closure is determined by the task-specific scientific requirements needed to resolve the research question, rather than by a threshold on the Atomic Score. The rubric links designated atomic criteria to closure requirements, but each requirement is assessed against its specified scientific conditions independently of the credit awarded for the corresponding criterion. Unmet indispensable requirements constrain the attainable RC level according to their scientific role. The resulting RC0--RC4 states follow the semantics summarized in Table~\ref{tab:rc_levels}. Appendix~\ref{app:worked_case30} presents selected criteria from grading rubric of Case~30.

\subsection{Human Consistency Check of Selected Grading Decisions}

Five physics-trained experts reviewed 100 evaluation cards covering all 50 benchmark cases. For each case, two submissions from different models were selected, with two material rubric criteria examined per submission, yielding 200 requested criterion-level judgments. The review cards contained the candidate evidence, relevant physical context, the selected rubric criteria, and accepted equivalent routes. Criteria were selected to provide broad case coverage and to emphasize scientifically consequential, locally assessable grading decisions rather than to form an equal-probability sample of all rubric atoms.

Of the 200 requested judgments, 192 yielded effective decisions, with 81.8\% of the judgments agreeing with the corresponding automated criterion-level decisions. RC-relevant consistency was assessed using atomic criteria linked to essential task-specific requirements, including central scientific objects, indispensable components, principal mechanisms, and material validation. These roles concern scientific requirements whose failure can limit the attainable RC level. The RC-relevant comparisons covered 84 judgments across 72 cards and 44 tasks, with 60 agreements among 79 binary judgments (75.9\%). Both rates measure criterion-level agreement with automated atomic judgments; RC assignment additionally accounts for task-specific trigger conditions and evidence dependencies. 

\section{Statistical Analysis}
\label{app:statistical_analysis}

The analysis comprises 1,050 final scored attempts originating from seven models, 50 cases, and three independent runs per model--case pair. The attempt is the basic analysis unit.
Incomplete attempts, including runs that reach the time limit, remain part of the scored population; infrastructure-invalid executions and superseded retries do not create additional observations.

For attempt $a$ and dimension $d\in\{F,O,I,M\}$, let $s_{ad}$ and $n_{ad}$ denote satisfied and applicable atoms. Functional completion is
\[
x_{ad}=\frac{s_{ad}}{n_{ad}},\qquad
\bar{x}_{Gd}=\frac{1}{|G|}\sum_{a\in G}x_{ad},
\]
where $G$ denotes the attempts belonging to the model or Research Closure group being summarized. N/A atoms enter neither the numerator nor the denominator. Terminal-zero attempts without atom-level grading contribute $x_{ad}=0$. 
The reported group value is the mean of these attempt-level completion rates, so each attempt contributes equally regardless of its number of applicable atoms.

Validation/Scope (V) is summarized separately at the criterion level as the fraction of applicable V criteria that are satisfied within the reported group. Applicability depends on the method used, the evidence produced, and the resulting conclusion. N/A criteria and terminal-zero attempts without atom records do not enter this criterion-level denominator.

Besides, in the joint score--RC distribution of Figure~\ref{fig:closure_association}A, each cell is a percentage of all 1,050 attempts, not an RC proportion conditional on its score band.

\section{Additional Results}
\label{app:additional_results}

\subsection{Detailed Dimension and Research Closure Results}

Tables~\ref{tab:atomic_dimension_model_support}--\ref{tab:rc_distribution_round} report the model-level and run-level values underlying the main-text dimension and Research Closure summaries. F/O/I/M follow the attempt-level convention defined in Appendix~\ref{app:statistical_analysis}; V is reported separately using its applicable-criterion denominator. The RC tables report all five closure levels. Run-level summaries cover the three independent runs for each model--case pair.

\begin{table}[H]
\centering
\caption{Functional completion by model. F/O/I/M values are mean attempt-level completion rates, with terminal-zero attempts contributing zero. V values represent the percentage of applicable V criteria satisfied, excluding N/A criteria and terminal-zero attempts without atom records.}
\label{tab:atomic_dimension_model_support}
\small
\begin{tabular}{@{}lrrrrrl@{}}
\toprule
Model & F (\%) & O (\%) & I (\%) & M (\%) & V (\%) & V satisfied/applicable \\
\midrule
Claude Opus 5 & 85.6 & 60.1 & 54.6 & 52.8 & 84.0 & 263/313 \\
GPT-5.6 Sol & 83.8 & 60.5 & 47.9 & 46.0 & 74.2 & 241/325 \\
Kimi K3 & 77.4 & 49.7 & 43.4 & 40.9 & 68.4 & 214/313 \\
DeepSeek-V4.1-Flash & 67.2 & 37.3 & 35.2 & 35.7 & 71.3 & 214/300 \\
Qwen3.8-Max & 62.3 & 28.7 & 24.1 & 25.9 & 54.7 & 156/285 \\
GLM-5.3-Flash & 56.7 & 23.7 & 20.3 & 23.3 & 65.6 & 177/270 \\
Gemini 3.1 Pro & 61.9 & 22.7 & 13.7 & 15.3 & 47.8 & 121/253 \\
\bottomrule
\end{tabular}
\end{table}

\begin{table}[H]
\centering
\caption{Functional completion by models in three independent runs. Measures and applicability conventions follow Table~\ref{tab:atomic_dimension_model_support}.}
\label{tab:atomic_dimension_round_support}
\scriptsize
\begin{tabular}{@{}llrrrrrl@{}}
\toprule
Model & Run & F & O & I & M & V & V satisfied/applicable \\
\midrule
Claude Opus 5 & R1 & 88.2 & 62.5 & 58.3 & 56.2 & 84.1 & 90/107 \\
 & R2 & 83.7 & 63.6 & 56.8 & 55.5 & 82.5 & 85/103 \\
 & R3 & 84.8 & 54.4 & 48.5 & 46.9 & 85.4 & 88/103 \\
\addlinespace[1pt]
GPT-5.6 Sol & R1 & 86.8 & 58.7 & 49.3 & 48.3 & 76.8 & 86/112 \\
 & R2 & 78.5 & 57.0 & 39.0 & 39.6 & 66.3 & 67/101 \\
 & R3 & 86.2 & 65.9 & 55.3 & 50.0 & 78.6 & 88/112 \\
\addlinespace[1pt]
Kimi K3 & R1 & 74.8 & 43.0 & 35.7 & 34.0 & 59.6 & 65/109 \\
 & R2 & 81.0 & 54.9 & 51.8 & 46.5 & 68.2 & 75/110 \\
 & R3 & 76.3 & 51.3 & 42.7 & 42.2 & 78.7 & 74/94 \\
\addlinespace[1pt]
DeepSeek-V4.1-Flash & R1 & 65.0 & 36.5 & 34.2 & 33.9 & 68.9 & 71/103 \\
 & R2 & 71.8 & 39.1 & 36.8 & 38.3 & 69.2 & 74/107 \\
 & R3 & 64.8 & 36.2 & 34.7 & 35.0 & 76.7 & 69/90 \\
\addlinespace[1pt]
Qwen3.8-Max & R1 & 55.5 & 25.1 & 17.8 & 17.9 & 41.5 & 34/82 \\
 & R2 & 66.8 & 31.4 & 28.2 & 32.2 & 65.0 & 65/100 \\
 & R3 & 64.7 & 29.7 & 26.2 & 27.5 & 55.3 & 57/103 \\
\addlinespace[1pt]
GLM-5.3-Flash & R1 & 58.8 & 31.7 & 27.2 & 28.0 & 62.6 & 62/99 \\
 & R2 & 61.2 & 25.9 & 21.7 & 25.2 & 57.9 & 55/95 \\
 & R3 & 50.2 & 13.6 & 12.2 & 16.6 & 78.9 & 60/76 \\
\addlinespace[1pt]
Gemini 3.1 Pro & R1 & 66.5 & 22.5 & 15.0 & 14.2 & 36.2 & 38/105 \\
 & R2 & 60.2 & 22.6 & 14.5 & 17.1 & 57.3 & 43/75 \\
 & R3 & 59.2 & 23.2 & 11.5 & 14.5 & 54.8 & 40/73 \\
\bottomrule
\end{tabular}
\par\vspace{2pt}\footnotesize F--V entries are percentages; the final column reports the numbers of satisfied and applicable V criteria.
\end{table}

\begin{table}[H]
\centering
\caption{Research Closure distribution by model. Cells report count (percentage) over the 150 final scored attempts for each model.}
\label{tab:rc_distribution_model}
\scriptsize
\begin{tabular}{@{}lrrrrrr@{}}
\toprule
Model & N & RC0 & RC1 & RC2 & RC3 & RC4 \\
\midrule
Claude Opus 5 & 150 & 24 (16.0\%) & 49 (32.7\%) & 25 (16.7\%) & 3 (2.0\%) & 49 (32.7\%) \\
GPT-5.6 Sol & 150 & 28 (18.7\%) & 38 (25.3\%) & 43 (28.7\%) & 7 (4.7\%) & 34 (22.7\%) \\
Kimi K3 & 150 & 48 (32.0\%) & 36 (24.0\%) & 32 (21.3\%) & 5 (3.3\%) & 29 (19.3\%) \\
DeepSeek-V4.1-Flash & 150 & 75 (50.0\%) & 25 (16.7\%) & 30 (20.0\%) & 4 (2.7\%) & 16 (10.7\%) \\
Qwen3.8-Max & 150 & 76 (50.7\%) & 39 (26.0\%) & 22 (14.7\%) & 2 (1.3\%) & 11 (7.3\%) \\
GLM-5.3-Flash & 150 & 87 (58.0\%) & 32 (21.3\%) & 19 (12.7\%) & 4 (2.7\%) & 8 (5.3\%) \\
Gemini 3.1 Pro & 150 & 76 (50.7\%) & 31 (20.7\%) & 41 (27.3\%) & 1 (0.7\%) & 1 (0.7\%) \\
\bottomrule
\end{tabular}
\end{table}

\begin{table}[H]
\centering
\caption{Research Closure counts by model and independent run. Each row reports 50 scored attempts.}
\label{tab:rc_distribution_round}
\scriptsize
\begin{tabular}{@{}llrrrrrr@{}}
\toprule
Model & Run & N & RC0 & RC1 & RC2 & RC3 & RC4 \\
\midrule
Claude Opus 5 & R1 & 50 & 8 & 14 & 7 & 1 & 20 \\
 & R2 & 50 & 11 & 9 & 12 & 2 & 16 \\
 & R3 & 50 & 5 & 26 & 6 & 0 & 13 \\
\addlinespace[1pt]
GPT-5.6 Sol & R1 & 50 & 7 & 13 & 14 & 5 & 11 \\
 & R2 & 50 & 15 & 9 & 13 & 2 & 11 \\
 & R3 & 50 & 6 & 16 & 16 & 0 & 12 \\
\addlinespace[1pt]
Kimi K3 & R1 & 50 & 20 & 13 & 7 & 2 & 8 \\
 & R2 & 50 & 14 & 10 & 15 & 1 & 10 \\
 & R3 & 50 & 14 & 13 & 10 & 2 & 11 \\
\addlinespace[1pt]
DeepSeek-V4.1-Flash & R1 & 50 & 30 & 5 & 9 & 1 & 5 \\
 & R2 & 50 & 22 & 12 & 6 & 3 & 7 \\
 & R3 & 50 & 23 & 8 & 15 & 0 & 4 \\
\addlinespace[1pt]
Qwen3.8-Max & R1 & 50 & 27 & 14 & 6 & 0 & 3 \\
 & R2 & 50 & 26 & 12 & 6 & 1 & 5 \\
 & R3 & 50 & 23 & 13 & 10 & 1 & 3 \\
\addlinespace[1pt]
GLM-5.3-Flash & R1 & 50 & 28 & 8 & 8 & 2 & 4 \\
 & R2 & 50 & 24 & 15 & 6 & 2 & 3 \\
 & R3 & 50 & 35 & 9 & 5 & 0 & 1 \\
\addlinespace[1pt]
Gemini 3.1 Pro & R1 & 50 & 25 & 11 & 13 & 0 & 1 \\
 & R2 & 50 & 23 & 13 & 14 & 0 & 0 \\
 & R3 & 50 & 28 & 7 & 14 & 1 & 0 \\
\bottomrule
\end{tabular}
\end{table}

\paragraph{Validation/Scope categories.}
Table~\ref{tab:validation_scope_categories} reports the detailed definitions for different V categories and their criterion-level satisfaction across the benchmark.  N/A criteria and terminal-zero attempts without atom records are excluded from the applicable-criterion denominator.

\begin{table}[H]
    \centering
    \caption{Validation/Scope criteria across the benchmark.}
    \label{tab:validation_scope_categories}
    \begin{tabularx}{\linewidth}{@{}lXrr@{}}
    \toprule
    V category & Description & Satisfied/applicable & Rate \\
    \midrule
    Scope justification &
    Justify the conditions under which the conclusions are valid. &
    852/981 & 86.9\% \\
    Required validation &
    Perform validation checks required by the task, regardless of the chosen method. &
    93/156 & 59.6\% \\
    Risk-linked validation &
    Address specific risks arising from the method, results, or conclusions. &
    441/922 & 47.8\% \\
    \bottomrule
    \end{tabularx}
\end{table}

All 26 RC3 attempts retain an unmet applicable V criterion, while 25 also satisfy another V criterion.

\FloatBarrier

\subsection{Atomic-score Distributions across Investigations}
\label{app:score_variability}

\begin{figure}[tbp]
    \centering
    \includegraphics[width=\linewidth]{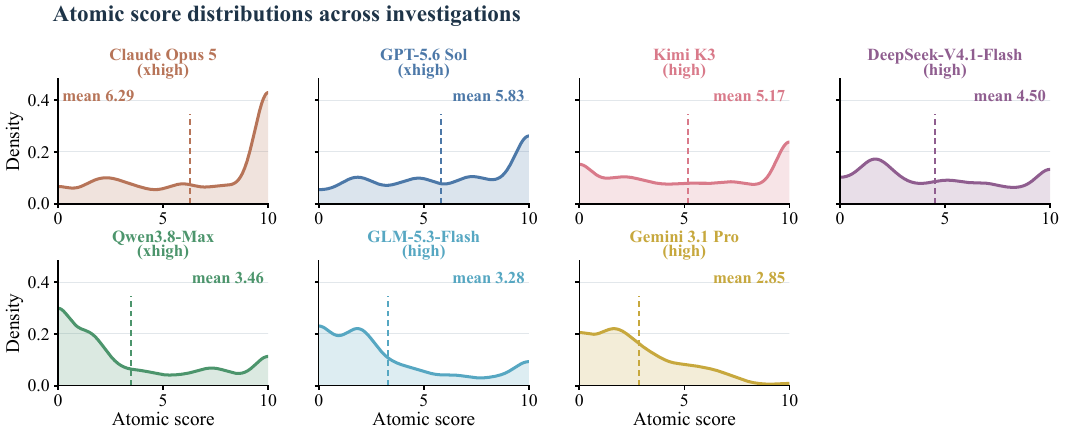}
    \caption{\textbf{Atomic-score distributions across investigations.}
    Kernel density estimates show the distribution of Atomic Scores over all final scored attempts for each model, including terminal-zero outcomes. Dashed lines indicate mean Atomic Scores for each model. The distributions summarize score variation across the full set of benchmark investigations.}
    \label{fig:score_variability_supp}
\end{figure}

Figure~\ref{fig:score_variability_supp} complements the aggregate model scores by showing the full distribution of Atomic Scores across each model's 150 scored attempts. These distributions reflect variation across benchmark tasks as well as across independent runs of the same task.

\FloatBarrier

\section{Representative Evidence Paths}
\label{app:representative_trajectories}

Figure~\ref{fig:representative_pathways} compares the scientific requirements supported by three Case~30 submissions. Table~\ref{tab:case30_research_activities} summarizes key research activities; the paragraphs below give the decisive evidence for each closure outcome.

The task asks when post-release central-potential scattering can account for an intensity-dependent attoclock offset, and whether the scattering-based explanation for a Coulomb potential also applies to screened potentials with the same binding energy. Answering the question requires calculating the offset, identifying and explaining this change, and establishing the interpretation's applicability. Functionally equivalent validity arguments are permitted rather than requiring a unique formula or validation route.

\begin{table}[H]
\centering
\caption{Selected Case~30 research activities and key findings. The steps summarize recorded activities, not a complete or strictly ordered execution history. Scores and RC levels refer to final submissions.}
\label{tab:case30_research_activities}
\begin{tabularx}{\linewidth}{@{}>{\raggedright\arraybackslash}p{0.17\linewidth}>{\raggedright\arraybackslash}p{0.44\linewidth}>{\raggedright\arraybackslash}X@{}}
\toprule
Investigation & Selected research steps & Key finding or unresolved issue \\
\midrule
Kimi K3 \newline 10.0/RC4 &
Constructed scattering calculations; detected an angle-sign inconsistency; corrected and checked the results; interpreted the screening dependence. &
Supported the interpretation within a physically justified scope. \\
\addlinespace
GPT-5.6 Sol \newline 9.5/RC3 &
Repaired the numerical solver; identified screening-dependent trends; explained their mechanism; recorded a high-field diagnostic. &
Did not resolve the applicability concern revealed by its high-field diagnostic. \\
\addlinespace
Gemini 3.1 Pro \newline 4.5/RC2 &
Revised trajectory propagation and angle convention; generated Coulomb and screened results; summarized the intensity dependence. &
Misidentified increasing or turning-over screened curves as monotonically decreasing. \\
\bottomrule
\end{tabularx}
\end{table}

\paragraph{Kimi: correcting the angle and establishing applicability.}
Kimi initially obtained opposite offset signs from an orbit integral and direct trajectory propagation. Correcting the angle convention reduced their discrepancy to $4.53\times10^{-7}$~rad, checking numerical consistency. It interpreted the screening dependence through the relative sizes of the tunnel-exit distance and screening length. Separately, the report restricted physical interpretation to quasistatic, sub-barrier exits and labeled other regimes as formal extensions. Two saved intensities fall within its stated Coulomb window, approximately $0.0065<I<0.0078$~a.u.; both retain a pronounced screening contrast. These restrictions support the reduced-model conclusion.

\paragraph{GPT: an unresolved high-field applicability risk.}
GPT correctly identified and explained the screening-dependent offset changes by balancing a shorter tunnel-exit distance against increasing electron momentum as intensity rises, but a separate issue arose in the high-field regime. At $E_0=0.08$~a.u., its saved diagnostic flags a field above the Coulomb-reference barrier-suppression threshold, $E_{\mathrm{BSI}}=0.0625$~a.u. This is not a universal threshold for screened targets. Nevertheless, the final report called the $\gamma\leq1$ regime more favorable for its weak-field exit approximation. A low Keldysh parameter does not resolve the separate barrier-suppression risk. The mechanism is supported, but the report did not adequately reconcile this diagnostic with its high-field interpretation, yielding RC3.

\begin{figure}[H]
\centering
\includegraphics[width=0.68\linewidth]{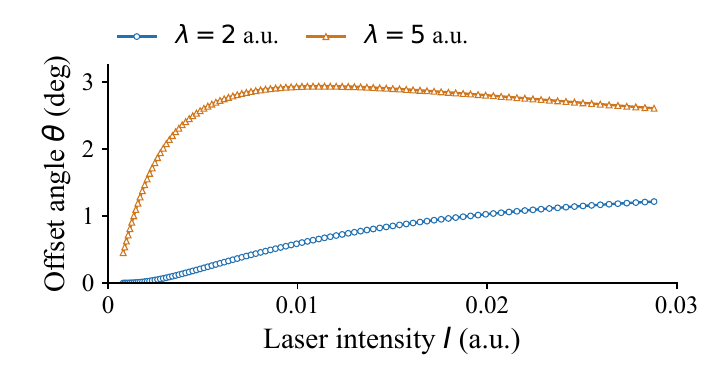}
\caption{Screened offsets contradict Gemini's decreasing-trend claim. Replotted from the agent's submitted numerical data. All 100 recorded points per curve are shown, with straight connectors and no fitting or smoothing. The angle $\theta$ is the submitted outgoing-trajectory offset in degrees, with $I=2E_0^2$; $\lambda$ is in atomic units.}
\label{fig:case30_gemini_data}
\end{figure}

\Needspace{0.17\textheight}
\paragraph{Gemini: an intensity trend contradicted by its own data.}
Gemini generated offset results for Coulomb and screened potentials, but incorrectly summarized their intensity dependence. Its report states that the calculated offset angle $\theta$ decreases monotonically with increasing laser intensity for all tested potentials. The saved $\lambda=1$--$3$ curves instead rise, while $\lambda=5$ and $10$ rise and then fall. Figure~\ref{fig:case30_gemini_data} shows their full submitted scans for $\lambda=2$ and $5$. Gemini discussed screening suppression and classical timing limits, but this misidentification prevented its evidence from supporting the screening-dependent response and its mechanism, yielding RC2.

\section{Supplementary Model-Level Results}
\label{app:reallocated_detail}

Table~\ref{tab:model_summary} provides a consolidated view of model-level Atomic Scores, attempt-level RC4 rates, and task-level RC4 consistency across the 50 benchmark tasks.

\begin{table}[H]
\centering
\caption{Model-level performance and RC4 consistency. Mean and median Atomic Scores and attempt-level RC4 rates are computed over 150 final scored attempts per model. The two RC4 consistency measures are computed over the same 50-task denominator for every model.}
\label{tab:model_summary}
\begin{tabular}{@{}lrrrrr@{}}
\toprule
Model & Mean score & Median & RC4 & $\geq$1 RC4 & All 3 RC4 \\
\midrule
Claude Opus 5 & 6.29 & 6.75 & 32.7\% & 50.0\% & 18.0\% \\
GPT-5.6 Sol & 5.83 & 6.00 & 22.7\% & 40.0\% & 10.0\% \\
Kimi K3 & 5.17 & 5.00 & 19.3\% & 40.0\% & 4.0\% \\
DeepSeek-V4.1-Flash & 4.50 & 4.00 & 10.7\% & 24.0\% & 0.0\% \\
Qwen3.8-Max & 3.46 & 2.00 & 7.3\% & 14.0\% & 2.0\% \\
GLM-5.3-Flash & 3.28 & 2.25 & 5.3\% & 12.0\% & 0.0\% \\
Gemini 3.1 Pro & 2.85 & 2.50 & 0.7\% & 2.0\% & 0.0\% \\
\bottomrule
\end{tabular}
\end{table}

\FloatBarrier

\section{Case 30 Task Specification and Rubric Excerpt}
\label{app:worked_case30}

This appendix reproduces the Case 30 task specification provided to the evaluated agents. Its formatting has been adapted to \LaTeX, but the scientific content is unchanged. Selected atomic criteria and case-specific Research Closure requirements are presented afterward.

\subsection{Case 30 Task Specification}

\begingroup
\small

\paragraph{When can a scattering model account for an attoclock offset?}

\paragraph{Background.}
In an attoclock measurement, the maximum of the photoelectron momentum distribution is displaced from the direction expected from the laser vector potential at ionization. That displacement is often discussed as a timing observable, but the outgoing electron also scatters from the residual binding potential.

\paragraph{Research problem.}
For a short circularly polarized pulse, determine whether the intensity dependence of the offset angle can be accounted for by a post-release central-potential scattering model. Establish how far a conclusion obtained for an unscreened Coulomb target remains valid as the interaction is screened, and state what the resulting evidence does and does not establish about a universal interpretation of the attoclock offset.

Choose the derivation, numerical evidence, comparisons, and presentation needed to make that assessment scientifically convincing. The number and arrangement of figures are for the candidate to decide.

\paragraph{Physical inputs.}
Work in atomic units. Use a circular driver of angular frequency $\omega=0.057$ and the intensity convention
\begin{equation}
I=2E_0^2, \qquad A_0=\frac{E_0}{\omega},
\end{equation}
where $E_0$ and $A_0$ are the peak electric-field and vector-potential amplitudes. Study a one-electron target with ionization potential $I_p=0.5$ under both
\begin{equation}
V_C(r)=-\frac{1}{r}
\end{equation}
and the screened family
\begin{equation}
V_{\lambda}(r)=-\frac{Z_{\lambda}}{r}e^{-r/\lambda}.
\end{equation}
For every finite $\lambda$, determine $Z_{\lambda}$ so that the field-free ground-state binding energy remains $I_p=0.5$. Include the Coulomb limit, the strong-screening case $\lambda=1$, and enough intermediate screening lengths of the candidate's choice to test whether an inference from either endpoint transfers across the family.

\paragraph{Method and scope boundary.}
Build the reduced model from the standard weak-field Keldysh exit scale, the stated vector-potential momentum scale, and the outgoing half of a central-potential scattering trajectory. Derive and declare the resulting angle convention. Standard Rutherford and screened central-potential scattering theory may be derived or cited. State the approximation level of the screened calculation used.

Choose an intensity domain that resolves the regimes needed for the claim while remaining consistent with the physical assumptions invoked. Numerical grids, root solvers, quadrature, sampling, tolerances, plotting ranges, and reliability tests are candidate choices. They are not grading targets in themselves; their consequences for the physical evidence are.

High-cost TDSE/iSURF propagation and CTMC regeneration are outside the required task. They may be discussed as external tests, but neither is required, and unsupported numerical data from those methods must not be presented as the candidate's own result.

\paragraph{Submission and resources.}
Submit one self-contained scientific report and an auditable offline calculation package. Include the task-specific derivation, code, numerical data underlying material claims, a provenance note listing every offline source or software document actually consulted and the method basis for items supplied from general knowledge, and concise environment and run information. No fixed file tree, filename scheme, figure count, or complete exploration history is required.

The accompanying resource contract supplies the binding limits: eight CPU cores, 45 minutes of candidate-active wall time, no network access, a 64-GiB scratch allowance, and a separate ten-minute sealing reserve. CPU usage is recorded as an audit metric rather than imposed as a second hard ceiling.

\paragraph{Reference and integrity policy.}
The examination is conducted without network access or live literature retrieval. Candidates may rely on existing general scientific knowledge and on any offline materials or software documentation explicitly supplied with the examination. Every consulted source must be recorded. For formulas, methods, constants, or conventions used from general knowledge rather than a consulted source, the method basis, definition, value, units, and convention must be stated sufficiently to make the calculation auditable; citations must not be invented.

The hidden target article and its identifying metadata, figures or captions, supplementary material, author code or data, target-specific reproduction code, solution or scoring material, online solutions, and worked reproductions of the case must not be used. All submitted scientific figures and numerical data must be generated by the candidate's own implementation of the stated problem.

\endgroup

\paragraph{Common submission requirements.}
In addition to the case-specific task instructions, candidates received benchmark-wide submission requirements for six top-level entries under \texttt{submission/}: \texttt{REPORT.md}, \texttt{DERIVATION.md}, \texttt{RUN\_LOG.md}, \texttt{REFERENCES.md}, \texttt{INTEGRITY.md}, and \texttt{ARTIFACTS.json}. These entries standardized submission collection, while the organization and naming of task-specific scientific files remained at the candidate's discretion.

\Needspace{0.38\textheight}
\subsection{Selected atomic criteria}

The Case 30 grading rubric contains twenty binary half-point atoms. Table~\ref{tab:case30_rubric_excerpt} presents representative criteria spanning the five functional dimensions. Research Closure is assessed separately using the case-specific requirements summarized in the following subsection.

\begin{table}[H]
\centering
\caption{Representative Case 30 atomic criteria. Criteria are condensed from the full case-specific grading rubric for readability.}
\label{tab:case30_rubric_excerpt}
\footnotesize
\setlength{\tabcolsep}{4pt}
\renewcommand{\arraystretch}{1.12}
\begin{tabularx}{\linewidth}{@{}p{0.08\linewidth}p{0.18\linewidth}X@{}}
\toprule
Atom & Function & Representative criterion \\
\midrule
F1 & Formulation & Relate the outgoing half of the central-potential deflection to the attoclock offset, using consistent exit and field/vector-potential conventions, units, and approximation limits. \\
O2 & Output & Calculate auditable offset-versus-intensity results for Coulomb and screened potentials matched to the same binding energy, including $\lambda=1$ and enough intermediate screening lengths to resolve changes with potential range. \\
I2 & Identification & Identify the approach to Coulomb behavior at long range, changes in intensity dependence at intermediate screening lengths, and the suppression and relative offsets at strong screening supported by the data. \\
M2 & Mechanism & Explain how the electron's exit or impact distance, asymptotic kinetic energy, and potential range compete to change the offset response as screening length decreases. \\
V1 & Validation/Scope & When numerical choices, binding-energy calibration, sampling, or curve structure raise a concrete risk to the conclusion, test that risk or restrict the conclusion accordingly. If no risk is triggered, record N/A without claiming positive validation evidence. \\
V3 & Validation/Scope & Specify which conclusions the reduced scattering calculation supports, distinguishing them from unverified real-atom, quantum-dynamical, or universal tunnelling-time interpretations. \\
\bottomrule
\end{tabularx}
\end{table}

\subsection{Case-specific Research Closure summary}

Research Closure for Case 30 is assessed separately from the atomic total, using the task-specific scientific requirements needed to answer the research problem. The requirements include a consistent offset calculation, comparisons of equally bound Coulomb and screened potentials, an explanation of the screening-dependent intensity response, and justified validation and applicability limits. The following levels illustrate Case~30, rather than a single decision template for all benchmark tasks:

\begin{list}{}{%
  \settowidth{\labelwidth}{\textbf{RC4:}}%
  \setlength{\labelsep}{0.6em}%
  \setlength{\leftmargin}{\labelwidth}%
  \addtolength{\leftmargin}{\labelsep}%
  \setlength{\itemindent}{0pt}%
  \renewcommand{\makelabel}[1]{\textbf{#1}\hfil}%
}
\item[RC0:] the central offset calculation or its mapping from outgoing-half scattering is missing or invalid;

\item[RC1:] the central offset is established, but the required comparison with screened potentials at the same binding energy is missing or invalid;

\item[RC2:] the required calculations exist, but the change in intensity dependence with screening, its mechanism, or another principal physical conclusion is unsupported or incorrect;

\item[RC3:] the principal results and mechanism are supported, but a concrete numerical-validation, physical-applicability, or interpretation-scope issue remains unresolved;

\item[RC4:] all indispensable scientific requirements and applicable validation/scope obligations are satisfied. An N/A risk-triggered Validation/Scope atom neither blocks RC4 nor constitutes positive validation evidence.
\end{list}

\FloatBarrier
\section{Grading Instructions}\label{app:grading_instructions}
\begingroup
\setlength{\parskip}{3pt}

\subsection{Grading Inputs and Assessment Stages}
\label{grading:inputs}

Each grading stage uses the common grading rules, the complete
task-specific rubric, and the following materials, with access
to previous assessments determined by the stage.

\subsubsection*{Required inputs}
\begin{itemize}
    \item \textbf{Research task:}
    The task specification, resource limits, and common
    submission requirements provided to the agent.

    \item \textbf{Task-specific grading rubric:}
    The atomic criteria, Research Closure requirements,
    applicable validation conditions, scientific dependencies,
    and accepted equivalent evidence.

    \item \textbf{Submitted scientific work:}
    The agent's research report, derivations, figures,
    code, numerical data, and supporting artifacts.

    \item \textbf{Execution records:}
    Records used to verify submitted results and determine
    whether a run is valid or requires re-execution.

    \item \textbf{Reference material:}
    Relevant scientific figures, passages, or validated
    reproductions used to check submitted results.
    These materials cannot supply scientific analyses
    or interpretations missing from the agent's work.
\end{itemize}

\subsubsection*{Independence and access}
The first-pass grader evaluates each submission without
access to previous assessments or other submissions.
The targeted reviewer can examine the first-pass assessment
and the supporting evidence.
If further adjudication is needed, the adjudicator receives
both earlier assessments and the remaining disputed issues.
At no stage may previous grades or aggregate results substitute
for scientific evidence from the submission.

\subsection{Shared Grading Instructions}\label{grading:common}

The following shared instructions apply to all three grading stages.

\begin{gradingprompt}{Shared Grading Instructions}

\subsubsection*{Scientific evidence and attribution}
Use the administered task, resource contract, complete task-specific rubric, and submitted scientific work. Do not change the rubric, repair the submission, or impose a requirement found only in the reference study. Do not use other submissions, prior scores, or aggregate results to determine the grade.

Distinguish four kinds of evidence: a scientific object exists in the submitted data (\textbf{E0}); the agent presents that object as evidence (\textbf{E1}); the agent identifies the scientifically important signal (\textbf{E2}); and the agent explains the physical inference supported by it (\textbf{E3}). These are different scientific acts. An evaluator-generated plot, slice, fit, or diagnostic can verify object existence and correctness, but cannot create omitted presentation, identification, mechanism, or validation evidence. Re-rendering can check an existing agent-authored presentation; it cannot replace one that the agent omitted.

Inspect every agent-authored figure supporting a scientific claim. Inspect code, data, and execution records when needed for a material scientific or validity decision. Assertions, proposed calculations, and unsubmitted exploratory results do not establish completed scientific work. Execution records can clarify validity or responsibility for a failure, but do not supply missing scientific reasoning. Separately authorized process evidence remains limited to its explicitly permitted scope.

\subsubsection*{Atomic scoring}
Each task contains twenty criteria, each worth either 0 or 0.5 points. Atomic Score is their sum on a 0--10 scale. Do not interpolate or invent fractional credit. A criterion is satisfied when its essential scientific requirement is met; an immaterial imperfection alone does not remove credit. A missing or incorrect essential requirement receives no credit. Record a material unresolved issue for review rather than assigning invented partial credit.

Formulation (F), Output (O), Identification (I), Mechanism (M), and Validation/Scope (V) denote scientific functions, not a required sequence of research actions. Apply their task-specific criteria and allocation. Aggregate score, code length, number of figures, and computational effort do not establish functional completion.

Mechanism credit requires the agent's presented signal and physical reasoning to support the inference. General knowledge alone, an unsupported correct conclusion, or figures connected only by the evaluator's explanation are insufficient.

For each Output criterion, separately check whether the scientific object is correct and reconstructible, whether the agent's presentation exposes its relevant structure, and whether the claim stays within what the evidence supports. Record these as \path{object_valid}, \path{presentation_adequate}, and \path{claim_adequate}. Output credit cannot be awarded if any condition is false. Check the principal structure, overall physical pattern, and scientifically important secondary structure; one superficially matching feature does not rescue a materially incorrect result.

\subsubsection*{Research Closure}

Assess Research Closure (RC0--RC4) using the complete
task-specific rubric, including the scientific requirements
for each level and their necessary dependencies.
Determine RC independently of the Atomic Score, dimension
totals, or individual criterion credits.
Complete closure requires the agent's work to satisfy all
indispensable scientific requirements, including the evidence,
interpretation, and applicable validation and scope conditions
needed to support the final conclusion.

Assess each closure condition against its specified scientific facts, independently of the corresponding criterion's credit. Inadequate presentation or an unsupported claim does not by itself establish that an otherwise valid required object is absent or physically invalid. When several closure failures apply, use the most severe applicable level specified by the rubric. Explain the essential unmet requirement, or why all indispensable requirements are satisfied. RC imposes no score ceiling, floor, multiplier, or preferred distribution.

\subsubsection*{Scientific dependencies and equivalent evidence}
Apply only scientifically necessary dependencies specified by the task-specific rubric. For a downstream failure, identify the exact failed prerequisite, the dependent requirement, and why the agent's evidence cannot establish that requirement independently. An unrelated missing part of a composite upstream criterion does not erase independently valid work. Refer a dependency without a necessary scientific premise to targeted review instead of silently changing the rubric. Do not penalize the same defect again as generic incompleteness or validation failure.

Accept scientifically valid alternative methods and representations within the rubric's equivalence conditions. Where relevant, compare physical identity, units, conventions, support, symmetry, sign, ordering, overall pattern, and component relations. Exact pixels, plotting style, color maps, reference grids, and unspecified absolute scales are not grading targets. If evidence cannot distinguish an acceptable equivalent from a substantive error, record the unresolved issue rather than inventing a numerical tolerance. Report an unauthorized rubric requirement or conflict with an acceptable equivalent as a rubric or fairness issue.

\subsubsection*{Validation and applicability}
Apply conditional criteria only when their task-specific conditions hold. In an assessable attempt, an inapplicable criterion is recorded as \path{not_applicable} (N/A) and retains its fixed 0.5-point contribution. It enters neither numerator nor denominator of applicable-criterion functional completion. An untriggered validation obligation is not positive evidence that validation was performed.

A valid run with no assessable scientific work receives score 0 and RC0 before criterion-level summation. N/A credit cannot create a score for such a run. Grade incomplete submissions and assessable timeouts normally; timeout alone does not make them zero. Infrastructure-invalid executions are excluded from scoring and rerun to obtain a valid measurement, rather than counted as agent failures. Other requests for a new execution require justification under the attempt-validity rules, not an automatic retry because a score is low.

Validation concerns a concrete scientific threat applying to the submitted method, evidence, or conclusion. Accept discriminating evidence or an evidence-based scope restriction that addresses that threat; do not require one preferred check. Generic refinement, residuals, or caveats do not automatically earn or lose validation credit.

Before assigning RC3 for a validation or scope gap, establish all six requirements:

\begin{enumerate}
\item \textbf{Failure mode:} identify the concrete failure mode that remains unexcluded.
\item \textbf{Origin:} explain why it arises from the agent's method, data, or conclusion.
\item \textbf{Recognizable risk:} identify the fair basis in the task or standard scientific reasoning for recognizing it.
\item \textbf{Affected conclusion:} identify the central conclusion that the failure mode threatens.
\item \textbf{Insufficient evidence:} explain why the submitted evidence is insufficient to expose or exclude this failure mode. Recognizing a risk does not resolve it without discriminating evidence or an evidence-based restriction of the threatened conclusion.
\item \textbf{Equivalent remedies:} give examples of functionally equivalent evidence that would resolve the issue, without requiring a unique validation method.
\end{enumerate}
Without these elements, do not assign RC3 by default. Restricting a conclusion to an established domain removes unsupported breadth; it does not repair an incorrect signal, an absent indispensable component, or a missing principal mechanism.

\subsubsection*{Verification of important results}
Consult reference evidence when a material score or RC decision depends on feature resolution, physical versus numerical fine structure, required components and their relations, formula identity, or representational equivalence. This includes full Output credit dependent on a nonstandard projection, a cropped range, or reconstruction from submitted data. Prefer applicable reference figures or passages, then validated reproductions within their established scope, and source pages if needed. Resolve insufficient or conflicting evidence using the relevant original source; this reading order does not authorize an out-of-scope reference to override applicable scientific evidence.

Before comparison, establish the observable, case and component, units, basis, gauge, normalization, and coordinate or transform conventions, as applicable.

A provisional Atomic Score of at least 9.0 \textbf{or} provisional RC4 requires a focused check of one central output and one principal-mechanism link. Check physical identity, sign, ordering, and overall pattern as applicable. Internal consistency and numerical convergence alone are insufficient. The reviewer verifies that these checks are applicable and support the result, and performs any missing necessary check before acceptance.

Record the submitted evidence, reference location, representation mapping, findings, scientific basis, affected decision, and remaining uncertainty. Distinguish confirmation from a changed judgment. Reference verification checks submitted results; it cannot complete scientific reasoning the agent omitted.

\end{gradingprompt}

\subsection{Independent First-Pass Grading Prompt}\label{grading:first}

\begin{gradingprompt}{Independent First-Pass Grading Prompt}

\subsubsection*{Role and inputs}
Grade one submitted investigation independently using the \hyperref[grading:inputs]{grading inputs (Section~\ref{grading:inputs})}, complete task-specific rubric, and \hyperref[grading:common]{shared grading instructions (Section~\ref{grading:common})}. Inspect the materials read-only; do not manage the examination, repair the submission, compare agents, inspect prior grades, or change the rubric. Use only this attempt's permissible evidence.

\subsubsection*{Scientific decisions}
First establish attempt validity and whether assessable scientific work exists. Apply the terminal-zero rule before criterion-level scoring; otherwise grade the available work, including partial work.

For an assessable submission:

\begin{enumerate}
\item Decide each of the exact twenty criteria as \path{pass}, \path{fail}, \path{unresolved}, or \path{not_applicable}, applying the rubric's scientific and applicability conditions.
\item Record all three Output conditions and complete the rubric's central scientific checks.
\item Apply the shared evidence-attribution, necessary-dependency, and equivalent-evidence rules. Preserve independently valid work without supplying missing presentation, identification, reasoning, or validation.
\item Assess RC independently from criterion credit, using the most severe applicable unmet closure requirement or explaining why all indispensable requirements are satisfied.
\item Record unresolved fairness issues and any justified request for a new execution under the attempt-validity rules.
\end{enumerate}
Inspect every agent-authored figure supporting a scientific claim and conduct the required reference checks, including the central-output and principal-mechanism checks when the high-result trigger applies.

\subsubsection*{Consistency checks}
Before returning the assessment, verify three conditions:

\begin{enumerate}
\item An Output criterion cannot pass when \path{object_valid}, \path{presentation_adequate}, or \path{claim_adequate} is false.
\item A downstream failure follows an absent or incorrect necessary scientific premise, not an unrelated display defect or missing component. Refer contradictory dependencies to targeted review.
\item RC conditions follow their specified scientific facts independently of criterion credit. Failed presentation or claim adequacy alone does not establish absence or invalidity of an existing valid object or component.
\end{enumerate}
Provide a short evidence-based reason for failed or unresolved criteria and central checks. Passing criteria do not require repetitive justifications.

\subsubsection*{Output}

Return a single structured JSON assessment recording the run's
validity, decisions on all twenty atomic criteria, the required
Output checks, and Research Closure with its supporting rationale.
Provide brief evidence-based reasons for failed or unresolved
criteria. Record the agent-authored figures inspected and any
reference checks performed, noting whether those checks support
or contradict the submitted results. Include any unresolved
issues requiring further review.

This is the first-pass assessment, not the accepted final result. Contradictory Output flags and credit, evaluator-created presentation, unsupported RC, or a missing or inadequate mandatory check prevent unchanged acceptance at review.

\end{gradingprompt}

\subsection{Targeted Review Prompt}\label{grading:review}

\begin{gradingprompt}{Targeted Review Prompt}

\subsubsection*{Role and inputs}
Review the first-pass assessment using the \hyperref[grading:inputs]{grading inputs (Section~\ref{grading:inputs})}, complete task-specific rubric, and \hyperref[grading:common]{shared grading instructions (Section~\ref{grading:common})}. You may inspect the initial assessment. This is a targeted review, not a routine second independent blind grade. Do not rerun calculations, repair the submission, change the rubric, or overwrite the initial assessment.

Verify submission identity, attempt validity, and scientific assessability. Apply the terminal-zero and partial-work rules before reviewing criterion credit.

\subsubsection*{Material checks}
Check decisions affecting Atomic Score, RC, fairness, or attempt validity. Verify physical-object and formula identity; inspection of all claim-supporting figures; scientifically important output structure; evidence attribution; necessary dependencies; accepted equivalents; and the distinction between mechanism and validation. Check task-specific closure conditions and all six elements supporting any RC3 finding.

Independently verify the \hyperref[grading:first]{first-pass consistency checks (Section~\ref{grading:first})}, including agreement between Output credit and its three conditions, justified necessary dependencies, and RC decisions independent of criterion credit. Preserve independently valid work and reject repeated deductions for the same defect.

For each high-result trigger in the shared instructions, verify that the first pass completed an applicable central-output check and principal-mechanism check. Perform any missing or inadequate required verification before acceptance. Prior grading records may identify a consistency issue but cannot replace the agent's evidence or establish a target score distribution.

\subsubsection*{Review decisions}
Start with the first-pass evidence map and disputed criteria. Reinspect code, data, reference material, or the complete rubric only as needed for a material issue; do not routinely repeat uncontested checks.

Record changed decisions, supporting submitted evidence, the applicable rule, relevant verification findings, and remaining uncertainty. Identify which first-pass findings are accepted.

Accept an assessment with no unresolved material issue, recording any corrections. Request focused adjudication only when disagreement could affect criterion decisions, RC, fairness, attempt validity, or whether a new execution is warranted. Do not average grades, impose RC-based score ceilings, or adjust results to an expected distribution.

\end{gradingprompt}

\subsection{Focused Adjudication Procedure}\label{grading:adjudication}

Adjudication addresses only material disagreements remaining after targeted review, not a routine third full assessment. The adjudicator receives the task, complete rubric, relevant submitted evidence, first-pass assessment, review decisions, and precise disputed question. The \hyperref[grading:common]{shared grading instructions (Section~\ref{grading:common})} apply throughout.

\subsubsection*{Decision procedure}
\begin{enumerate}
\item \textbf{Identify the dispute and affected requirements.} Verify the attempt and evidence identity, the exact disputed criterion or closure condition, and its basis in the task. Retain the attempt's validity classification unless validity itself is disputed. Identify unauthorized requirements or equivalence conflicts as rubric or fairness issues.
\item \textbf{Examine the relevant evidence.} Inspect the affected submitted figures, artifacts, and permissible reference material. Apply the shared attribution, equivalence, comparison, and high-result verification rules. Complete any missing mandatory check without rerunning calculations, repairing the submission, or supplying missing agent reasoning.
\item \textbf{Resolve the affected decisions.} Apply the shared atomic, applicability, necessary-dependency, and RC rules only to decisions affected by the dispute. Preserve unrelated valid work and uncontested decisions. If an acceptable equivalent cannot be distinguished from a substantive error, retain the uncertainty rather than inventing a definite judgment.
\item \textbf{Record the decision and rationale.} State the dispute, whether it is resolved, submitted and reference evidence examined, applicable rule, scientific reasoning, changes from review, and resulting score or RC where determined. Explain any effect on fairness, attempt validity, or a justified request for a new execution.
\end{enumerate}
The targeted reviewer integrates the adjudication with the first-pass and review assessments. Preserve the initial assessment and record changes explicitly; do not average judgments. Prior assessments identify the dispute, not the scientific answer. Other submissions and expected score distributions do not determine the decision.

\subsection{Grading Output Format}
\label{grading:output}

The first-pass grader produces a structured assessment recording
the validity of the run, decisions on the twenty atomic criteria,
checks of key scientific results, and Research Closure with its
supporting rationale. The assessment also records relevant evidence,
reasons for failed or unresolved decisions, and any issues requiring
further review.

The Atomic Score is calculated from the accepted criterion-level
decisions using the prescribed half-point and N/A rules. Runs with
no assessable scientific work receive zero credit. Targeted review
and, when necessary, adjudication document any changes to the
initial assessment and the evidence supporting those changes.

\endgroup

\end{document}